\documentclass[10pt,twocolumn,letterpaper]{article}
\usepackage{cvprlike}
\usepackage[pagebackref=false,breaklinks=true,colorlinks=true,
            linkcolor=black,citecolor=black,urlcolor=black]{hyperref}

\newcommand{\ie}{\textit{i.e.}}

\title{Explaining AI-Image Detection: What the Heatmap Actually Shows}

\author{%
Leonid Kuturin\textsuperscript{1} \quad
Ilya Sotnikov\textsuperscript{1} \quad
Mark Khusnutdinov\textsuperscript{1} \quad
Mikhail Potemkin\textsuperscript{1} \\
Pavel Baranas\textsuperscript{1} \quad
Aleksandra Korepanova\textsuperscript{1} \quad
Alexander Kalashnikov\textsuperscript{2} \\[0.5em]
{\normalsize \textsuperscript{1}Sirius Educational Centre, Big Challenges programme \\
\textsuperscript{2}HSE University, Faculty of Computer Science}%
}

\begin{document}
\maketitle

\begin{abstract}
A marketplace review photograph is a document: platforms approve refunds on it, and
generative models drove the cost of forging one to zero. We study that detection problem,
so we build a detector and attach an attribution map as its evidence. We
measure what that pair delivers on 186{,}527 images, under controls designed to
change our conclusions when something is wrong. Our scope is fully generated photographs,
and locally edited ones enter as a diagnostic. The detector we selected fires on 142 of 368
local edits, against 19 for the first-fix one, the model our first repair produced. We find
that compression history, not synthesis, drives naive evaluation. Trained on the corpus as
collected, our strongest model reaches 0.9999 PR-AUC (the area under the precision--recall
curve) on a product-disjoint split. It falls to 0.7254 once we re-encode synthetics into
the real class's format, while five public detectors move by at most 0.07. We find that
aligning one class relocates the cue rather than removing it. The repaired model then
assigns native user files a median probability of synthesis of 0.0004. One identical final
encode for both classes repairs that, and in a three-seed factorial we credit the encoding
change with the whole gain ($+0.176 \pm 0.009$ PR-AUC).
That encode equalises the last stage only, so we read the rest. Hand-designed forensic
features alone still separate the classes at 0.7145 against a base rate of 0.254. Across
eight encoding histories our selected model's median image moves 0.0001 of the probability
range, while 33.0\% of images still cross its threshold under some history.
For evidence we test maps causally: deleting the highlighted region must move the score.
Whether an attribution ranking exists at all depends on whether the detector reacts to
the image. On the first-fix detector, which calls 96 of 100 edited frames real, we measure
no advantage for any map over a random one. On the detector we selected we re-measure
seventeen maps against that control on both axes at once. Twelve clear it on edited
images and eight on generated ones. We find three facts on both axes: perturbation leads,
attention rollout is third, and no gradient-CAM variant shows a positive advantage. The
controls that never consult the detector never clear it either, and on generated images
the centre prior is worse than random. Our ensembled regional map clears both axes and
takes the top pixel AP, per-pixel average precision against the edited region. It is level
with the centre prior, at 12.4 seconds per map against 44.9 for occlusion. An interval covering
zero does not certify emptiness, and clearing a detector-blind control is not yet a
faithful explanation. Each control is cheap, and each one changed a conclusion.
\end{abstract}

\begin{figure}[t]
  \centering
  \includegraphics[width=\columnwidth]{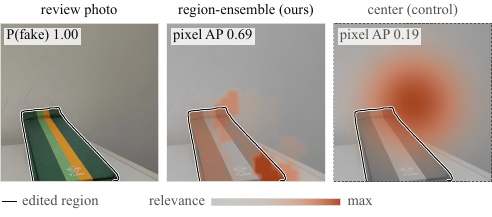}
  \caption{A marketplace review photo with a local AI edit. The outline is the region the editing tool was asked to change. Middle: \texttt{region\_ensemble}, the map we selected, puts its relevance on the edited object. Right: the \texttt{center} prior never consults the detector and misses here, yet priors like it win pooled localisation metrics on centre-biased benchmarks. Chips: $P(\text{fake})$, the detector's probability of synthesis for the photo, then each map's pixel AP, per-pixel average precision against the outline, recomputed from the arrays drawn.}
  \label{fig:teaser}
\end{figure}

\section{Introduction}
\label{sec:intro}

A photograph attached to a marketplace review is a document. Buyers decide on it,
moderators approve refunds on it, and platforms pay for it. Each of those roles is an
incentive to forge it, and generative models made forging free. We study the detection of
such forgeries among genuine review photographs. A rejection has to be justified on
appeal, so we ask whether a post-hoc attribution map can serve as evidence beside the
verdict. Both halves are hard to measure here. The two classes reach a corpus by different
routes, so a detector can score perfectly on provenance alone. The evidence for synthesis
is a statistic spread over the frame rather than a compact object. A map that outlines the
right thing therefore need not be reporting a cause. We deliver a measured account of that
pair on marketplace review photography.

We build the argument in two stages. In Stage 1 we build the detector from what is already
known and report what it does. In Stage 2 we fix what would count as an explanation of it,
then report whether anything meets that bar. Off-the-shelf parts cover the first stage
only partly. Detectors lose recall on generator families they were not trained on
\cite{wang2020cnndetection,ojha2023universal,corvi2023diffusion}. And a detector answers
with a single number, which a moderator cannot show and an appeal cannot cite.

On this task both of our metrics turn into self-deception. The detector metric is fooled
by \emph{shortcuts}: artefacts that correlate with the label but carry no trace of
generation. Our two classes differ in file format and compression history. A model can
therefore reach a near-perfect held-out score, on a split whose products never appear in
training, without detecting synthesis at all \cite{grommelt2024fakeorjpeg}. The
explanation metric is fooled differently. A map can be smooth, reproducible and convincing
while having nothing to do with what the model responds to \cite{adebayo2018sanity}.

We therefore fixed the rules of measurement before quoting any number. We read the
detector on three axes. An axis is one fixed image set read under one fixed condition ---
here the compression histories recompressed, symmetric, and native. We accept an
explanation only if its paired advantage over the best trivial control has a lower bound
above zero. Trivial controls are maps that never consult the detector. The bound comes
from a cluster bootstrap that resamples whole scenes rather than single images, wherever
the run covers the whole mask release. The check on a map is causal: confidence must drop
when we remove the highlighted region. Both headline results change under these rules, so
the controls are our subject.

\paragraph{What is new here.} That codec provenance drives detection scores is known
\cite{grommelt2024fakeorjpeg,ricker2026biasid}, and we claim no discovery of format bias.
Four things are ours. We show that an asymmetric repair \emph{relocates} the shortcut
rather than removing it. Our first model reads 0.9999 PR-AUC on held-out data and 0.7254
once we re-encode the synthetic class, while five public detectors shift by at most 0.07.
On a native-file axis we measure what that relocation costs on the
distribution a detector would receive. With a three-seed factorial we attribute the gain to
one of the two changes we made together. And we apply the same falsification discipline to
the detector and to its explanations, reporting where each fails. We find the attribution
ranking to be a property of the detector. A bundle here is a frozen release of encoder,
features, head and threshold. On one that calls almost every edited frame real, nothing
separates from a random map; on the one we selected, twelve of seventeen maps do
\cite{tomsett2020saliencymetrics}.

\paragraph{What this paper does not show.} We do not demonstrate a faithful explanation.
Our acceptance rule is necessary and not sufficient: a map that beats a detector-blind
control has passed one causal test, under one replacement strategy, at one sample size.
Failing that rule is likewise a failure to demonstrate superiority at this protocol and
sample size, not a proof that a method is empty. The ranking it produces is conditional
on the detector, as we show on two bundles. We close only the provenance channel the final
encode creates. And we run no moderator study, so nothing here says whether a map helps a
human decide.

\section{Related Work}
\label{sec:related}

Two literatures bound this work: what detectors read, and whether a map can be audited.
Detectors exploit the stable low-level trace a generator leaves
\cite{wang2020cnndetection}, and that trace transfers poorly to diffusion models
\cite{corvi2023diffusion}. The datasets under those numbers are the problem. Common
corpora store real images as JPEG and generated ones as PNG, so a detector partly solves
``guess the format''
\cite{grommelt2024fakeorjpeg,zhu2023genimage,ricker2026biasid,zhou2026fragile}. Dual data
alignment matches the classes in pixel and frequency space so the label cannot be read off
the pipeline \cite{chen2025dda}. We inherit that diagnosis and ask what happens after the
obvious repair.

The measures that audit post-hoc maps are fragile in their own right, to imperceptible
input changes and to arbitrary design choices
\cite{adebayo2018sanity,ghorbani2019fragile,kindermans2019unreliability}. That is enough
to make comparative rankings untrustworthy \cite{tomsett2020saliencymetrics}. Others run
that comparison with a random baseline and multiplicity correction, on object-centric
classification, and several methods do clear it \cite{skliarov2025comparative}. Deleting a
region to test a map is also suspect, since the deleted image leaves the data distribution
\cite{hooker2019roar,rong2022road}. The paired counterfactual operator instead replaces
it with the pixels of a source image \cite{goyal2019counterfactual,eckstein2021discattr}.
Weakly-supervised localisation was in turn largely explained by a centre prior. That is
why a protocol scores against the best trivial control rather than a random map
\cite{choe2020wsoleval}, as saliency benchmarks have long done
\cite{tatler2007centralbias,bylinskii2019metrics}. We inherit that rule rather than
propose it. A parallel field trains localisers that predict edited pixels directly
\cite{guillaro2023trufor,tantaru2024weakloc,huang2025sida,zhou2025aigiholmes}. We train
none: our model stays a whole-image binary classifier, explained after the fact.

\section{Stage 1: Building the Detector}
\label{sec:stage1}

This stage contains no discoveries. We take the frozen backbone, the crop geometry and
the feature blocks from established practice, and argue each choice rather than claim it.
One thing did change under us: the training data.

\subsection{Corpus and split}
\label{sec:data}

Four constraints shape the corpus: genuine user photography, several generator families,
a product-disjoint split, and an identical compression history for both classes.
Product-disjoint means no product appears on both sides of the split. As collected, our
corpus meets the first three and fails the last. The real class is 169{,}751 photographs
from public marketplace product reviews. We store them at full size, and they arrive
\emph{already in WebP} because the service that hosted them re-encoded them into that
format. We discard a frame that contains a face rather than
blurring it, since a blur is itself an artefact correlated with the class. The synthetic
class is 16{,}776 images from 10 generators served through OpenRouter, one reference and
one composite prompt per image, stored as \emph{JPEG q85}. Labels come from generation
records, so they carry provenance rather than human annotation. Each generation inherits
the product group of its reference. We cut by product group $80/20$ into 149{,}215
training and 37{,}312 held-out test images, at a fake share of 8.9\% and 9.2\%. The
$1{:}10.1$ imbalance leaves the synthetic class rare, so our primary metric is PR-AUC.

We report two numbers that answer different questions, and we keep them apart. R@P95
throughout this paper is recall read off the \emph{test} precision--recall curve where
precision first reaches 95\%. It needs no threshold and is not an operating-point number.
We select every decision threshold on validation only and transfer it unchanged. We
label a rate that depends on it ``at the validation threshold''. Metrics whose threshold
we tune on test labels appear nowhere here.

\subsection{What the model sees}
\label{sec:detector}

One architectural decision matters. We freeze the backbone and train only the head.
13{,}350 synthetic training images across ten generators do not constrain a large ViT, so
full fine-tuning would fit those generators rather than synthesis. Frozen features are
also the stronger reported baseline here
\cite{ojha2023universal,cozzolino2024raisingbar}.
We ran no end-to-end fine-tuning, so we argue the choice rather than measure it.

The descriptor concatenates three blocks. The first is the pooled pre-logit embedding of a
frozen encoder. The second is a radially binned log spectrum of the grey central crop
\emph{at native resolution}. The third is residual statistics from the fixed high-pass
kernels of the spatial rich model (SRM). We compare DINOv2
\cite{oquab2023dinov2}, CLIP \cite{radford2021clip}, SigLIP \cite{zhai2023siglip},
DINOv3 and Perception Encoder \cite{bolya2025pe} in the first block. The other two give
44 \emph{forensic} features, i.e.\ hand-designed statistics of the pixel grid rather than
learned semantics. The head is a small MLP, stopped early on validation PR-AUC, with the
threshold and a Platt calibration, one logistic rescaling of the logit, fitted on
validation \cite{guo2017calibration}. The
single-run detector comparison below holds \texttt{seed}~$=42$ across configurations. The
factorial uses three seeds per cell, and other runs use the seed their job recorded.

Geometry decides what each branch can sense. The processor resizes the shorter edge to 336
and centre-crops a square, so on a non-square frame the model sees only the central band.
We compute every map and localisation metric here inside that crop. The forensic blocks
instead run on the native crop and sense compression history directly, while the visual
branch is nearly blind to codec traces. Switching them off therefore treats the format
shortcut rather than merely ablating features, and we call that configuration
\emph{visual-only}.

\subsection{Compression history as a leakage channel}
\label{sec:axes}

Reals arrive as WebP and synthetics as JPEG, so a naive build hands the model an almost
perfect separating feature. We therefore measure the same model under three fixed
compression histories, three \emph{axes} in the sense of Sec.~\ref{sec:intro}. \textbf{v2}
re-encodes the synthetic class to WebP. \textbf{v2-webp2} applies one final encode to
both classes, LANCZOS to 1200\,px then WebP q80. The \textbf{native control} keeps
synthetics in their original JPEG. We call the first recipe \emph{asymmetric} and the
second \emph{symmetric}, after whether the same pixel operation reaches both classes. We
designed a symmetric JPEG control and never ran it, because it could create a
new shortcut of its own.

The native axis needs one caveat, and we state it here. Its two sides carry different
encoding histories. Real frames were already re-encoded by the service that hosted them,
while synthetic files are exactly as the generator returned them. That is what a detector
would receive at upload: a mixed-provenance axis rather than one clean
distribution. It is decisive because it is the axis a detector would meet in use, and
every native number here carries that mixture.

\subsection{The shortcut, and what repaired it}
\label{sec:shortcut}

\paragraph{An alarm, not a result.}
We read our best held-out score as an alarm.
Every run here pairs a frozen backbone and forensic features with a light head at
\texttt{seed}~$=42$, read on the held-out test split of Sec.~\ref{sec:data}.
PE-Core-336 \cite{bolya2025pe} reaches \textbf{0.9999} PR-AUC at R@P95 \textbf{1.000}, on
a corpus taken as collected.
We then re-encoded the synthetic side of the same test set to WebP q80, the format the
real frames are stored in.
The model drops to \textbf{0.7254} PR-AUC at \textbf{0.340} R@P95.
Most of the ``perfect'' score was format detection \cite{grommelt2024fakeorjpeg}.

With an external control we locate the effect in the data rather than in the
transformation.
We ran the same WebP test through five public synthetic-image detectors, each with its
own preprocessing.
Three are \texttt{haywoodsloan/\allowbreak ai-image-detector-deploy} (SwinV2),
\texttt{Ateeqq/\allowbreak ai-vs-human-image-detector} (SigLIP) and
\texttt{Smogy/\allowbreak SMOGY-Ai-images-detector} (Swin).
The other two are
\texttt{NYUAD-ComNets/\allowbreak NYUAD\_AI-generated\_images\_detector} (ViT) and
\texttt{Organika/sdxl-detector} (Swin).
We fetched each from its Hugging Face default branch on 2026-07-17, with no revision
pinned.
Their PR-AUC moves by at most 0.07, in both directions.
They are weak on our data to begin with, so from that control we conclude only that the
transformation is not destructive.
The collapse is a property of \emph{our} training data.
On the same test we see our own ranking invert: PE-Core-336 loses $-0.274$ and DINOv3-L
only $-0.032$.

\paragraph{First fix, and why it failed.}
Our first fix scored well on the metric and failed on native files.
We re-encoded the synthetic class with the same WebP q80.
Retrained PE-Core-336 reaches PR-AUC \textbf{0.9976} on the recompressed axis, and we call
that bundle the \emph{champion}, or the \emph{first-fix} bundle after the repair that
produced it.
Then we read it on files in the format they arrived in: \emph{the best model always says
no}.
The median $P(\text{fake})$ on those synthetic files was \textbf{0.0004}.
The model caught \textbf{24.7\%} of them at the validation threshold.
We had normalised the synthetic class only, while real frames kept the trace of their
earlier re-encode, so the shortcut had not been removed.
It had mutated from ``WebP vs.\ JPEG'' into ``our re-encoder vs.\ everything else''.

\paragraph{Second fix, and which half of it worked.}
The second fix changed two things at once.
We applied the same final WebP encode to \emph{both} classes, and we switched the
forensic features off, since they are the primary sensors of compression history.
On the decisive \emph{native} test the median native-fake probability moves from 0.0004
to 0.9999, and PR-AUC gains $+0.181$, from 0.6732 to \textbf{0.8540}.
The two recompressed numbers, 0.9976 asymmetric and 0.9792 symmetric, are each recipe's
own axis.
The $-0.018$ between them is a change of operating point rather than a controlled delta.

We therefore completed the $2\times2$ at three seeds per cell (Tab.~\ref{tab:factorial}).
One pooled standard error of \textbf{0.0094} serves all three effects.
A main effect is what one factor changes on average, with the other averaged over.
For symmetric encoding it is $+0.176 \pm 0.009$ native PR-AUC, an order of magnitude above
seed noise.
The main effect of dropping the forensic features is $-0.000 \pm 0.009$: on average, the
change we bundled with the fix did \emph{nothing}.
The two factors interact, at $+0.036 \pm 0.009$ on that convention.
Read as simple effects, the forensic branch costs 0.036 native PR-AUC while the shortcut
is open and buys 0.036 once both classes pass the same final encode.
The best cell is therefore not the one the second fix produced, which pays $-0.036$ for
the features it drops.
We adopt that cell as our final configuration.
Three seeds per cell estimate an interaction poorly, so we read its sign and order of
magnitude rather than its exact value.
We retrained ten of the twelve runs independently of our cluster, and they reproduce the
numbers we report.

\begin{table}[t]
\centering
\footnotesize
\setlength{\tabcolsep}{4pt}
\caption{The completed factorial, three seeds per cell, mean $\pm$ sample sd. PR-AUC on
the native test, the same 37{,}312-image set for all cells. In small type, PR-AUC on each
recipe's own recompressed axis --- different test sets, not comparable across cells. The
effects in the text are $\pm\frac{1}{2}$ contrasts over these four means, sharing one
pooled standard error of 0.0094.}
\label{tab:factorial}
\begin{tabular}{lcc}
\toprule
 & asym.\ \texttt{webp} & sym.\ \texttt{webp2} \\
\midrule
full (FFT$+$SRM) & $0.680 \pm 0.016$ {\scriptsize 0.998} & $\mathbf{0.892 \pm 0.007}$ {\scriptsize 0.984} \\
visual-only      & $0.716 \pm 0.028$ {\scriptsize 0.998} & $0.857 \pm 0.003$ {\scriptsize 0.980} \\
\bottomrule
\end{tabular}
\end{table}

\subsection{What the detector is, and what it still reads}
\label{sec:whatdetector}

Our selected detector is the winning cell of that factorial: a frozen PE-Core-336
encoder, the 44 forensic features, and a small MLP head over symmetrically encoded data.
It reads 0.892 native PR-AUC over three seeds.
Symmetric encoding equalises the last stage only, so we measured the rest.
We refit the same head on disjoint channels of one backbone pass, fitted on 42{,}991
images and read on a 10{,}359-image validation split at a synthetic base rate of 0.254.
On the symmetric corpus the 44 forensic features alone still reach \textbf{0.7145} PR-AUC.
That residual channel, what the container still tells the model after the symmetric
encode, is weakened rather than closed.
Our selected model barely uses it: adding the forensic block to the visual features moves
validation PR-AUC by $\mathbf{+0.0009}$.

Whether a model \emph{follows} the container is a separate question, so we put it to the
files.
We scored 1{,}500 real and 1{,}500 synthetic native test files under eight encoding
histories, from one WebP pass to JPEG q50.
That run is balanced at a 50\% base rate, so its PR-AUC is not on the scale of the
8.9\%-prevalence numbers elsewhere in this paper.
The first-fix bundle tracks the container.
Its median image sweeps \textbf{0.9988} of the unit interval across those histories, at an
interquartile range of [0.014, 1.000].
\textbf{59.0\%} of its images cross the decision threshold under some history, 97.5\% of
synthetics.
The selected model is much steadier and is not invariant.
Its median image moves \textbf{0.0001}, but the same distribution has an interquartile
range of [0.000, 0.990] and a mean of 0.328.
\textbf{33.0\%} of images still cross the threshold under some history, 59.1\% of
synthetics.
We do not call that invariance.
Assigning each file's history by a hash of its name, independent of the label, costs the
selected model nothing: 0.9769 PR-AUC, against 0.8459 for the first fix.
The model also has an honest ceiling, and it is generator coverage rather than
calibration.
Per-generator PR-AUC on the native axis falls to 0.366 on \texttt{mai\_image}.
Holding out a whole generator family, we read 0.979, 0.921 and 0.919 across three folds of
the selected recipe.

\section{Stage 2: Explaining the Detector}
\label{sec:stage2}

A verdict alone cannot be appealed, so we ask what a detector can \emph{show} a
moderator, stating the acceptance rule first and the results second.

\subsection{What would count as an explanation}
\label{sec:protocol}

We treat a map as a claim about causes, and a claim about causes has to survive an
intervention.
A map counts as an explanation here only if deleting what it highlights moves the
detector's score more than deleting what the best trivial control highlights.
The margin must also carry an interval that clears zero.

\paragraph{One contract, three trivial controls.}
We make every method explain one scalar, the detector's raw logit before sigmoid and
calibration, and reuse its feature extraction and head loading verbatim.
We record a method that cannot run as \texttt{invalid}, and never silently retarget it.
We registered 28 implementations and benchmark 18: three mandatory controls and fifteen
methods from the perturbation, attention, gradient, CAM and region families.
All share one patch geometry and one replacement strategy, i.e.\ whatever fills a region
once it is deleted, inside the crop the model sees.
The controls \texttt{random}, \texttt{center} and \texttt{edge} enter every run: uniform
noise, a centred blob, and the edge map of the image.
None consults the detector.
They are the bar, not a courtesy baseline.

\paragraph{The causal test, and the localisation anchor.}
\textbf{Faithfulness} asks whether deleting the highlighted region actually moves the
score \cite{petsiuk2018rise,deyoung2020eraser,yeh2019infidelity,wu2024saco}, and AOPC, the
area over the perturbation curve, is the mean score drop as ranked regions are removed in
order.
\textbf{Localisation} scores maps against reference intervention masks
\cite{zhang2018pointinggame}.
Pixel AP is average precision with every pixel as one decision.
RMA, or relevance mass, is the share of a map's positive relevance landing inside the
mask.
The pointing game asks whether a map's peak falls inside the target.
The anchor is 368 triples of background, edited image and mask, 183 object insertions and
185 inpaintings.
A mask marks the region a tool was asked to change, so the generator may leave traces
outside it and, in principle, none inside.
Recomputing from the pixels what each tool actually changed, we see how loose that anchor
is, and how unevenly.
Median leakage outside the requested mask is 34.5\% of the changed area for the 183
\texttt{gpt} triples, against 1.14\% for \texttt{flux} and 0.19\% for \texttt{bria}.
In 37 triples more of the realised change falls outside the mask than inside.
The pooled median of 17.5\% hides that split, so we name the generator behind every
localisation claim.

\paragraph{The aggregation unit, the interval, the rule.}
A near-duplicate background is not a second observation.
The set holds \textbf{217 distinct scenes behind 368 triples}; a scene is one background
photograph and every triple built on it.
The scene is therefore our primary bootstrap unit on a run that covers the whole release.
Every such interval averages per-unit means rather than rows, at 2000 resamples and
2.5/97.5 percentiles.
Switching from \texttt{base\_id}, one unit per edited frame, to scene there widens
intervals by 4.8--63.7\% and changes no ordering.
The unified run below is the one run that carries both axes.
It samples 200 of those triples and clusters by frame, so its intervals are of the
narrower kind, and we say so at its table.
A method counts as working only if the bootstrap lower bound of its per-image paired
advantage exceeds zero, against the best trivial control on that axis.
That control is \texttt{random} on faithfulness, and \texttt{center} or \texttt{edge} on
localisation, whichever is stronger on the run.
We read every method twice.
\emph{Conditionally}, we score only valid maps that carry positive energy, i.e.\ relevance
above zero somewhere in the map.
\emph{End-to-end}, every invalid map, empty map and detector miss scores a pre-registered
zero, fixed in code before any run.
One weakness is ours to declare in advance: our only replacement strategy is blur.

\subsection{Two families of our own}
\label{sec:ours}

Neither family is a new class of method. Both adapt to one property of this task: the
evidence is spread over the frame. The unit a method perturbs therefore matters more than
the estimator on top of it.

\paragraph{Pooling attribution.}
The head sits on attention pooling, so the pooled vector is a weighted sum of token
contributions, and both methods exploit that structure rather than the input.
\texttt{pool\_decomposition} scores a token by the inner product of its contribution to
the pooled vector with the logit gradient with respect to that vector, in one backward
pass.
\texttt{pool\_occlusion} instead ablates a token fully inside the pooling, replaying all
ablations in one batch.
Each map costs under 0.2\,s on the edited frames, which makes them usable at interactive
latency.
Neither localises well, because depth dilutes a single token's contribution.
And \texttt{pool\_decomposition} is exact only for the pooled vector, not for the logit a
non-linear head computes from it.

\paragraph{The region family.}
A region method bets that the evidence sits on a connected entity, a whole object or one
edited region, which the square token lattice tears apart.
We therefore move causal attribution onto SLIC superpixels \cite{achanta2012slic}, small
connected groups of similar pixels that follow object boundaries.
LIME introduced that idea \cite{ribeiro2016lime}; ours is the task-specific adaptation and
its measurement.
\texttt{object\_region} blurs one superpixel of \emph{the crop the model sees} at a time
and takes the logit drop as attribution.
\texttt{compact\_region} adapts extremal perturbations \cite{fong2019extremal}, the search
for the smallest region whose deletion still moves the score, under an area budget.
\texttt{patch\_anomaly} is a single-pass forensic map, the deviation of a patch embedding
from the mean.
\texttt{region\_ensemble} is the consensus of the three, snapped to the same superpixels.
Where each stands against the controls is the subject of Sec.~\ref{sec:xai-results}, and what
each leads on is metric-specific, so we say so at every claim.

\subsection{Whether any map beats the control depends on the detector}
\label{sec:xai-results}

\paragraph{Setup.}
We benchmark 18 maps on the detector we selected, and one method set carries both axes.
We score seventeen of them as an advantage over the eighteenth, the \texttt{random}
control.
The bundle is the winning factorial cell of Sec.~\ref{sec:whatdetector}: symmetrically encoded
data with the forensic features on.
The \emph{edits} axis is 200 locally edited frames and their reference masks, drawn at
sampling seed 42 from the 368-mask release, and this detector fires on 75 of them.
The \emph{generation} axis is 60 fully synthetic images.
Faithfulness and localisation come from that one run, on the same frames and the same
maps (Tab.~\ref{tab:methods}).
Both axes use the corrected crop geometry of Sec.~\ref{sec:detector}.

\begin{table*}[t]
\centering
\scriptsize
\setlength{\tabcolsep}{3pt}
\caption{One method set, both axes, one bundle: the detector we selected, trained on
symmetric \texttt{webp2} data with the forensic features on. AOPC advantage is the
per-frame paired advantage over the \texttt{random} map on that axis, with 95\%
cluster-bootstrap intervals over that axis's frames at 2000 resamples. Bold marks an
interval that excludes zero, including the \texttt{center} control's generation entry,
which excludes it from below. Localisation is conditional on a valid map with positive
energy, on the same 200 edited frames, and \texttt{ms/map} is that run's cost. Controls
in italics. $n_{\text{valid}}$ is 200 on edits and 60 on generation, except
\texttt{vit\_gradcam} at 193 and 51. For localisation it is 200 except \texttt{rise} (5),
\texttt{compact\_region} (199) and \texttt{vit\_gradcam} (193), so the \texttt{rise}
localisation row rests on five maps. \texttt{smoothgrad} and
\texttt{integrated\_gradients} ran as their own single process, because they exhaust the
GPU under sharding; their controls reproduce this run's to three decimals.}
\label{tab:methods}
\begin{tabular}{lccccccc}
\toprule
 & \multicolumn{2}{c}{AOPC advantage over \texttt{random}} & \multicolumn{4}{c}{Localisation, edits axis} \\
\cmidrule(lr){2-3}\cmidrule(lr){4-7}
Method & edits & generation & pixel AP & RMA & Pointing & ms/map \\
\midrule
\multicolumn{7}{l}{\emph{Trivial controls}} \\
\textit{random} & --- & --- & 0.170 {\scriptsize [0.150, 0.192]} & 0.168 & 0.075 & 41 \\
\textit{center} & $+0.037$ {\scriptsize [$-0.055$, $+0.098$]} & $\mathbf{-0.058}$ {\scriptsize [$-0.120$, $-0.002$]} & 0.462 {\scriptsize [0.420, 0.504]} & 0.216 & 0.560 & 39 \\
\textit{edge} & $+0.015$ {\scriptsize [$-0.009$, $+0.040$]} & $-0.037$ {\scriptsize [$-0.085$, $+0.005$]} & 0.380 {\scriptsize [0.345, 0.415]} & 0.252 & \textbf{0.650} & 295 \\
\midrule
\multicolumn{7}{l}{\emph{From the literature}} \\
\texttt{rise} & $\mathbf{+0.177}$ {\scriptsize [$+0.148$, $+0.208$]} & $\mathbf{+0.106}$ {\scriptsize [$+0.071$, $+0.148$]} & 0.459 {\scriptsize [0.295, 0.631]} & 0.088 & 0.800 & 15{,}008 \\
\texttt{occlusion} & $\mathbf{+0.176}$ {\scriptsize [$+0.139$, $+0.220$]} & $\mathbf{+0.113}$ {\scriptsize [$+0.062$, $+0.155$]} & 0.256 {\scriptsize [0.231, 0.280]} & \textbf{0.287} & 0.605 & 44{,}893 \\
\texttt{attention\_rollout} & $\mathbf{+0.159}$ {\scriptsize [$+0.126$, $+0.197$]} & $\mathbf{+0.078}$ {\scriptsize [$+0.056$, $+0.100$]} & 0.360 {\scriptsize [0.327, 0.391]} & 0.195 & 0.420 & 2{,}153 \\
\texttt{smoothgrad} & $\mathbf{+0.128}$ {\scriptsize [$+0.093$, $+0.176$]} & $\mathbf{+0.069}$ {\scriptsize [$+0.045$, $+0.096$]} & 0.197 {\scriptsize [0.181, 0.214]} & 0.226 & 0.300 & 9{,}249 \\
\texttt{integrated\_gradients} & $\mathbf{+0.119}$ {\scriptsize [$+0.088$, $+0.158$]} & $\mathbf{+0.069}$ {\scriptsize [$+0.048$, $+0.092$]} & 0.180 {\scriptsize [0.164, 0.198]} & 0.207 & 0.170 & 1{,}048 \\
\texttt{raw\_attention} & $\mathbf{+0.065}$ {\scriptsize [$+0.032$, $+0.103$]} & $-0.032$ {\scriptsize [$-0.073$, $+0.004$]} & 0.387 {\scriptsize [0.361, 0.413]} & 0.280 & 0.475 & 304 \\
\texttt{layercam} & $+0.025$ {\scriptsize [$-0.007$, $+0.074$]} & $+0.015$ {\scriptsize [$-0.005$, $+0.036$]} & 0.164 {\scriptsize [0.146, 0.185]} & 0.140 & 0.080 & 181 \\
\texttt{vit\_gradcam} & $+0.006$ {\scriptsize [$-0.034$, $+0.053$]} & $\mathbf{-0.042}$ {\scriptsize [$-0.077$, $-0.011$]} & 0.112 {\scriptsize [0.098, 0.127]} & 0.084 & 0.098 & 135 \\
\texttt{gradcam\_plusplus} & $-0.002$ {\scriptsize [$-0.027$, $+0.036$]} & $-0.016$ {\scriptsize [$-0.038$, $+0.007$]} & 0.183 {\scriptsize [0.159, 0.208]} & 0.161 & 0.205 & 165 \\
\midrule
\multicolumn{7}{l}{\emph{Ours}} \\
\texttt{region\_ensemble} & $\mathbf{+0.155}$ {\scriptsize [$+0.124$, $+0.193$]} & $\mathbf{+0.059}$ {\scriptsize [$+0.026$, $+0.089$]} & \textbf{0.466} {\scriptsize [0.425, 0.506]} & 0.226 & 0.440 & 12{,}422 \\
\texttt{object\_region} & $\mathbf{+0.135}$ {\scriptsize [$+0.105$, $+0.170$]} & $+0.037$ {\scriptsize [$-0.013$, $+0.076$]} & 0.424 {\scriptsize [0.385, 0.463]} & 0.266 & 0.445 & 5{,}303 \\
\texttt{compact\_region} & $\mathbf{+0.082}$ {\scriptsize [$+0.050$, $+0.117$]} & $+0.026$ {\scriptsize [$-0.002$, $+0.051$]} & 0.272 {\scriptsize [0.241, 0.305]} & 0.268 & 0.246 & 5{,}353 \\
\texttt{pool\_occlusion} & $\mathbf{+0.064}$ {\scriptsize [$+0.035$, $+0.101$]} & $\mathbf{+0.066}$ {\scriptsize [$+0.042$, $+0.093$]} & 0.174 {\scriptsize [0.156, 0.194]} & 0.203 & 0.275 & 166 \\
\texttt{patch\_anomaly} & $\mathbf{+0.051}$ {\scriptsize [$+0.029$, $+0.075$]} & $+0.005$ {\scriptsize [$-0.028$, $+0.038$]} & 0.326 {\scriptsize [0.302, 0.350]} & 0.191 & 0.405 & 109 \\
\texttt{pool\_decomposition} & $\mathbf{+0.045}$ {\scriptsize [$+0.019$, $+0.073$]} & $\mathbf{+0.052}$ {\scriptsize [$+0.024$, $+0.085$]} & 0.138 {\scriptsize [0.121, 0.156]} & 0.179 & 0.240 & 146 \\
\bottomrule
\end{tabular}
\end{table*}

\begin{figure*}[t]
  \centering
  \includegraphics[width=\textwidth]{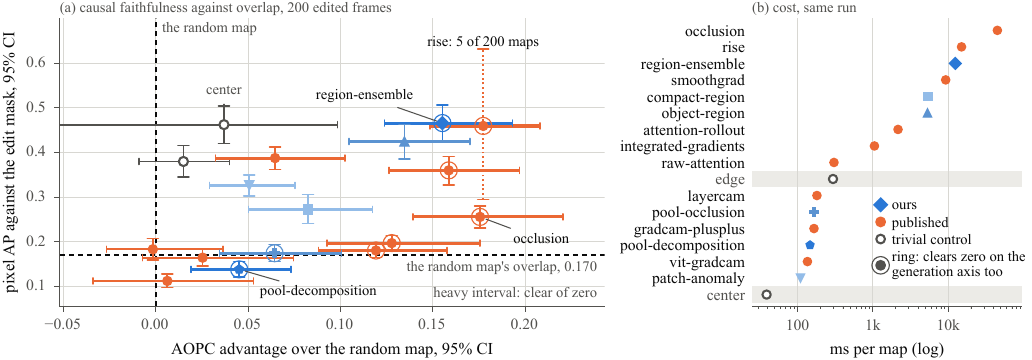}
  \caption{One bundle, one run, two metrics: the detector we selected, trained on symmetric \texttt{webp2} data with the forensic features on. (a) Seventeen maps on the same 200 locally edited frames. Across: AOPC advantage over the \texttt{random} map, the causal test. Up: pixel AP against the reference edit mask, overlap. Both carry 95\% cluster-bootstrap intervals, and a horizontal interval is heavy when it clears zero, which twelve do. The dashed rules are the \texttt{random} map itself, at zero advantage and 0.170 pixel AP. The quadrants are the finding: \texttt{occlusion} passes the causal test and lands far below the \texttt{center} prior on overlap, and \texttt{center} is second on overlap and never passes it. A ring marks a map that also clears zero from above on the generation axis, 60 synthetic images. \texttt{rise} is scored on the 5 frames whose map carries positive relevance, and we rank nothing by its dotted interval. Every other map is scored on all 200, except \texttt{compact\_region} (199) and \texttt{vit\_gradcam} (193). (b) What a map costs, in milliseconds. Cost has no interval, so it keeps its own panel.}
  \label{fig:axes}
\end{figure*}
\paragraph{The same seventeen maps, two detectors.}
We find that the headline moves with the detector and not with the protocol.
Our first-fix bundle calls 96 of 100 edited frames real.
On that bundle we measured no advantage for any map over the random one on the edits axis.
That is a statement about measuring maps of a nearly constant negative logit.
On the detector we selected, twelve of the seventeen maps clear the random control on the
edits axis and eight clear it on the generation axis (Fig.~\ref{fig:axes}).
Three facts hold on both.
Perturbation leads, with \texttt{rise} at $+0.177$ [$+0.148$, $+0.208$] on edits and
\texttt{occlusion} at $+0.113$ [$+0.062$, $+0.155$] on generation.
\texttt{attention\_rollout} is third on both, at $+0.159$ and $+0.078$.
And we measure no positive advantage for any gradient-CAM variant on either axis.
We therefore read the earlier negative result as conditional on the detector rather than
as a property of attribution.

\paragraph{The trivial controls are still the bar, and they now separate.}
Neither trivial control clears the random map on either axis.
On the generation axis \texttt{center} sits \emph{below} the random map with an interval
clear of zero, at $-0.058$ [$-0.120$, $-0.002$].
\texttt{vit\_gradcam} sits there too, at $-0.042$ [$-0.077$, $-0.011$].
Deleting the centre of a fully synthetic image moves the logit less than deleting a
random region of it.
We built the acceptance rule for exactly this comparison, since a method has to beat a
map that never consults the detector.
An interval that covers zero still says only that a method did not demonstrate
superiority at this protocol and sample size.
It does not certify that the method is empty.

\paragraph{Where our own methods stand.}
Our methods clear the control on the edits axis, and they do not lead it.
Of our region family only \texttt{region\_ensemble} clears both axes, for 12.4 seconds
per map; \texttt{object\_region} clears the edits axis for 5.3.
\texttt{pool\_occlusion} and \texttt{pool\_decomposition} clear both axes at 166 and 146
milliseconds, two orders of magnitude under an \texttt{occlusion} map.

\paragraph{One perturbation is not two.}
Our own rule asks for more than one perturbation, and this run supplies one.
We measured every advantage above under blur replacement, and blur is what
\texttt{occlusion} is built on.
On the first-fix bundle, switching the evaluator's replacement from blur to the channel
mean reversed the sign of the region family's advantage.
We did not repeat that control on the selected detector.
A faithfulness ranking is therefore a statement about a protocol as much as about a
method.

\subsection{Where the maps land}
\label{sec:loc-results}

\paragraph{The honest bar is still the best trivial control.}
We find a trivial spatial control winning the headline localisation metric on this run.
The \texttt{edge} control takes the pointing game with \textbf{0.650}, ahead of every
method that returns a map on all 200 frames, and \texttt{center} follows at 0.560
(Tab.~\ref{tab:methods}).
The best of those methods, \texttt{occlusion}, sits between them at 0.605.
That is a property of the dataset rather than of the methods.
Which control is best stays run-dependent, so we report both.

\paragraph{Overlap, on the run that measured faithfulness.}
No map leads on every localisation metric, and the leader changes with the metric.
By pixel AP \texttt{region\_ensemble} takes the top point estimate at \textbf{0.466}
[0.425, 0.506], with the \texttt{center} control at 0.462 [0.420, 0.504] and the random
map at 0.170 [0.150, 0.192].
The ensemble sits above the prior here, and the two intervals overlap almost entirely, so
we call them level.
By relevance mass \texttt{occlusion} leads at 0.287.
\texttt{raw\_attention} at 0.280, \texttt{compact\_region} at 0.268 and
\texttt{object\_region} at 0.266 also clear the best control, \texttt{edge} at 0.252.
By pointing the controls lead.
Faithfulness follows none of those orderings, in either direction.
\texttt{pool\_decomposition} clears the faithfulness control on both axes and scores 0.138
pixel AP, below the random map.

\paragraph{The prior's lead tracks edit size.}
A pooled pixel-AP comparison against the centre prior, our \texttt{center} control, is a
comparison at this set's mix of edit sizes.
Split by the protocol's pre-registered buckets, \texttt{region\_ensemble} reads 0.336 on
the 58 \texttt{small} frames against the prior's 0.290, and 0.743 on the 8 \texttt{xl}
frames against its 0.925.
A large edit is almost by construction a central one.
We print no interval per bucket and read the sign only.

\paragraph{What the pre-registered zeros do to this.}
Conditional numbers hide the detector's misses, so we read them end-to-end as well.
The selected detector fires on 75 of the 200 edited frames, and every miss takes a
pre-registered zero.
\texttt{region\_ensemble}'s pixel AP then falls from 0.466 to 0.200 [0.159, 0.242],
against 0.175 [0.137, 0.213] for \texttt{center} and 0.060 [0.045, 0.075] for the random
map.
That is a number about our system rather than about a map.
Re-scored over all 368 edits, the first-fix bundle fires on 19 against 142 for the one we
selected.
Its floor then collapses every method to 0.02--0.03, so the floor moves with the detector
and not with the maps.

\paragraph{We score one map on five frames, and one returns nothing on seven.}
\texttt{rise} leads the AOPC advantage on the edits axis and localises almost nowhere.
It returns a valid map on all 200 frames, and 5 of them carry positive relevance.
Its 0.459 pixel AP and its 0.800 pointing therefore rest on those five, and we rank
nothing by them.
RISE averages random masks weighted by the model's output, so a frame the detector calls
real offers no evidence for a mask to raise.
On the first-fix bundle that mechanism was exact, since the 96 maps without positive
energy were the same frames as the 96 detector misses.
Here the detector fires on 75 of 200 and only 5 maps carry positive energy, so saturation
explains part of it and not all.
\texttt{vit\_gradcam} fails in the other direction, returning an identically zero map on
7 of 200 frames, which our validity contract rejects as \texttt{blank\_map}.
\texttt{chefer\_relevance} \cite{chefer2021transformer} is the sharper warning, and it is
why this set is 18 maps and not 19.
On this attention implementation every layer's attention gradient comes back
\texttt{None}, so the method emits an identically zero map on every frame.
A standard method can produce nothing on a common stack, and only an explicit validity
contract catches it.

\paragraph{The map we would ship, measured.}
We arrived at our map in three steps, and the chronology matters because our own point
estimates chose the first.
We first chose \texttt{occlusion}, on its point-estimate lead in faithfulness.
Then came \texttt{object\_region} as a display variant, i.e.\ the same map computed on the
model's own tensor grid and sparsified before painting.
Our selected map is \texttt{region\_ensemble}, on that tensor grid with 48 SLIC
segments and a quantile-0.80 background floor, over the detector we selected.
A serving code path would return that variant, while the benchmark scores an image-space
implementation without the display floor.
We therefore compared like with like twice, over the same 368 masks.
Holding the detector at the first-fix bundle, the serving code path reaches 0.471 pixel
AP [0.437, 0.505] against 0.483 [0.450, 0.516] for the benchmark implementation.
The two are not distinguishable at this sample size.
Holding the detector at the selected one, the raw grid reaches 0.486 [0.449, 0.524] and
the display variant a viewer sees reaches 0.434 [0.401, 0.467].
The floor costs pixel AP and roughly doubles relevance mass, 0.476 against 0.278.

\section{Discussion and Limitations}
\label{sec:discussion}

\paragraph{A held-out split is no defence against shortcuts.}
Our split was correct and our detector still cheated.
Group splits protect against content leakage, not against file provenance.
Our real photos share one service's processing channel and our synthetics a generator
API's, identically in train and in test, so the channel itself is a label.
Removing one proxy only shifts the decision to the next, and symmetrising the final encode
closes the last stage only.
We do not close the residual channel.
We measure it, and report both what our selected model gains from it and how much its
verdict still moves.

\paragraph{An attribution ranking is a property of the detector.}
A map shown to a moderator is a claim about causes, so it must carry a causality
check.
We require the advantage over the best trivial control, with an interval, under more than
one perturbation.
On a bundle that calls 96 of 100 edited frames real, no method clears that rule, and that
is a fact about measuring maps of a saturated output.
On the bundle we selected, most of those same maps clear it.
Nothing about the methods changed between those two readings, so the ranking we can report
is a property of the detector.
The useful question is therefore not which method is faithful but which map is worth its
cost.
Inpainting masks inherit the bias of the process that made them.
And even with that bias cancelled, a map can put its peak inside the edit and spread its
relevance elsewhere.

\label{sec:limitations}
\paragraph{What bounds these conclusions.}
The localisation input is the weakest part of this work.
The mask set does not meet the specification we built it to.
Its lossy frames bound how precisely we can define an intervention region, and its leakage
is bimodal by generator.
Every localisation number is therefore a comparison between methods on one anchor, not an
absolute score.
The statistics are thin in places.
No ranking here is unconditional: our two replacement families disagree in sign where we
ran both.
The unified run also clusters by frame rather than by scene, which narrows its intervals
against the rest of this paper's.
Three seeds bound the factorial's interaction loosely, and 60 images bound the generation
axis.
We ran no human study, so a map's usefulness to a moderator is unmeasured.
Raw per-run artefacts live in a cluster store rather than in the repository, so from
source alone the procedure is verifiable and the result is not.

\section{Conclusion}
\label{sec:conclusion}

A careful product-disjoint split is not enough in this domain. It certified 0.9999 PR-AUC
for a model that was separating ``WebP vs.\ JPEG''. Diagnosis takes three axes, and the
cure takes two stages, because aligning only one class relocates the shortcut instead of
removing it. A factorial isolates the active ingredient: the symmetric encoding alone, not
the accompanying feature change. What survives the cure we measure rather than assume
away.

Explanations need the same discipline, and they need it applied to the right detector.
Measured on a bundle that calls almost every edited frame real, we find no method with an
advantage over a random map, because the maps describe a saturated output. On the detector
we selected we ran one method set over both axes. Twelve of seventeen maps clear that
control on edited images and eight on generated ones, and the ordering barely moves
between the two. Under a second replacement strategy we saw that advantage reverse sign
for our whole superpixel family. A faithfulness ranking is therefore a statement about a
protocol as much as about a method. Inpainting masks anchor localisation without human
annotation, but inherit a spatial prior whose strength tracks edit size. We find our
ensembled regional map level with the best trivial control on pixel AP, and ahead of every
other map scored on all 200 frames. The controls keep the pointing game.

Our final configuration is the winning cell of the factorial: a frozen Perception Encoder
with the forensic features on, trained on symmetrically encoded data. Beside it we would
return the display variant of \texttt{region\_ensemble}. We offer it as a localiser, not
an explanation. An empty map on a real photo is an honest answer.

{\small
\bibliographystyle{ieeetr}
\bibliography{refs}
}

\clearpage
\appendix
\renewcommand{\thesection}{\Alph{section}}
\setcounter{section}{0}
\setcounter{table}{0}
\setcounter{figure}{0}
\renewcommand{\thetable}{A\arabic{table}}
\renewcommand{\thefigure}{A\arabic{figure}}

\begin{center}
{\large\bfseries Appendix}
\end{center}
\vspace{0.5em}

\section{Extended Related Work}
\label{sec:supp-related}

Sec.~\ref{sec:related} states our positioning in short. This section is the full
discussion, and it cites the same work.

\paragraph{Synthetic image detection.}
Detectors exploit the stable low-level trace a generator leaves. That trace transfers
across convolutional families once training augments with compression and blur
\cite{wang2020cnndetection}. The transfer breaks for diffusion models, whose spectral
traces differ in kind \cite{corvi2023diffusion}, and that result reframed the field
around generalisation to unseen generators. A second shift replaced training from
scratch: light heads over frozen CLIP features generalise far better than fine-tuned
networks \cite{radford2021clip,ojha2023universal,cozzolino2024raisingbar}. We follow
that logic with a frozen ViT \cite{dosovitskiy2021vit} and a light head, comparing
SigLIP \cite{zhai2023siglip}, DINOv2 \cite{oquab2023dinov2} and PE-Core
\cite{bolya2025pe}. A decision threshold also needs calibration
\cite{guo2017calibration}. Adjacent work attributes the source model
\cite{sha2023defake} and benchmarks transfer at scale \cite{zhu2023genimage}.

\paragraph{Dataset bias and shortcuts.}
Common detection datasets store real images as JPEG and generated ones as PNG. A
detector then partly solves ``guess the format'' and loses much of its reported quality
once preprocessing is aligned \cite{grommelt2024fakeorjpeg}. Benchmarks also draw the
real class from existing corpora while generating the synthetic class through a separate
pipeline \cite{zhu2023genimage}. Two recent results sharpen the picture. A factorial transformation audit applies JPEG, WebP, resizing and rotation independently to each
class. It finds that codec provenance drives the scores of six published detectors
\cite{ricker2026biasid}. A controlled study of preprocessing shows per-generator
conclusions moving by up to 0.38 AUROC when the resize policy changes
\cite{zhou2026fragile}. On the training side, dual data alignment matches the two
classes in pixel and frequency space precisely so the label cannot be read off the
pipeline \cite{chen2025dda}.

Our contribution here is not the diagnosis but what happens next. The obvious remedy is to re-encode the synthetic class into the real class's container. That
does not remove the cue but relocates it, and the repaired model then goes quiet on the files
users actually upload. We measure that with an axis of native files, and we credit the repair to the encoding change
with a factorial. We then read the residual channel directly, by fitting the same head on the
forensic features alone. A shortcut-driven decision makes any attribution a
faithful map of the wrong cause, so shortcut control precedes explanation evaluation.

\paragraph{Attribution methods for vision transformers.}
The CAM family weights activation channels by gradients or by measured response changes
\cite{selvaraju2017gradcam,chattopadhay2018gradcampp,jiang2021layercam,wang2020scorecam,desai2020ablationcam}.
Porting it to a ViT is not automatic. Tokens must be re-arranged into a grid, and the
choice of layer, of normalisation and of CLS handling changes the map qualitatively.
Gradient methods act on the input directly
\cite{sundararajan2017ig,smilkov2017smoothgrad}. Perturbation methods need no internals
\cite{zeiler2014occlusion,petsiuk2018rise,ribeiro2016lime,lundberg2017shap}, at the
price of hundreds of forward passes. They operate naturally over superpixels
\cite{achanta2012slic,felzenszwalb2004graph,kirillov2023sam}, as our regional methods
do, and our compact variant follows extremal perturbations \cite{fong2019extremal}. A
separate branch reads attention
\cite{abnar2020rollout,chefer2021transformer,xie2023vitcx,bousselham2024legrad,wu2024tokentm}.
Most of this work targets semantic classification, where the concept sits in a compact
object. Evidence for synthesis is a distributed texture and spectrum statistic, so ``a
correct map outlines the object'' stops being a useful guide.

\paragraph{Evaluating explanations and localising manipulations.}
Maps are audited from several directions. Sanity tests randomise the weights
\cite{adebayo2018sanity}; intervention curves delete or insert the highlighted pixels
\cite{petsiuk2018rise,hooker2019roar}. Others measure infidelity and sensitivity
\cite{yeh2019infidelity}, comprehensiveness and sufficiency \cite{deyoung2020eraser},
ordering consistency \cite{wu2024saco} and the pointing game
\cite{zhang2018pointinggame}. These measures are themselves fragile to imperceptible input changes and to arbitrary design
choices \cite{ghorbani2019fragile,kindermans2019unreliability}. That unreliability is enough to
make comparative rankings between attribution methods untrustworthy
\cite{tomsett2020saliencymetrics}. Toolkits unify metric implementations, but not
aggregation, controls or statistics \cite{hedstrom2023quantus}; a meta-evaluation asks
which estimator to trust in the first place \cite{hedstrom2023metaquantus}. Where others run the comparison with a random baseline and multiplicity correction, on
object-centric classification, several methods do clear the baseline
\cite{skliarov2025comparative}.
Our negative result is therefore a statement about this task, not about attribution in general.
Evidence of synthesis is a distributed texture statistic rather than a compact object. Deleting a region to test it is itself suspect, since the deleted image leaves the data
distribution \cite{hooker2019roar,rong2022road}. Replacing the region with the pixels of a
paired source image avoids that. The operator was introduced for generating counterfactual
explanations \cite{goyal2019counterfactual} and later used to evaluate attribution
\cite{eckstein2021discattr}. We adopt it, with reference intervention masks
from real editing tools. Generative inpainting is the other way to stay in distribution \cite{cohen2025meaningfulpert}.
We cannot use it here: painting a region with a diffusion model injects the very trace the
detector looks for.

Trivial spatial priors as baselines are established practice. A centre prior was found to explain most of weakly-supervised localisation
\cite{choe2020wsoleval}. A protocol must therefore score against the best trivial control
rather than against a random map. Saliency benchmarks have long corrected for centre bias
\cite{tatler2007centralbias,bylinskii2019metrics,kummerer2018benchmarking}. We inherit
that rule rather than propose it.
Applied detection papers typically show a few maps as illustration, so the reader cannot
tell an informative map from a spatial prior. The exceptions below do report
localisation metrics, but for localisers of their own training, not for post-hoc maps of
a frozen classifier. A parallel field predicts edited pixels directly
\cite{guillaro2023trufor,guo2023hifiifdl,ma2023imlvit}, sometimes from JPEG artefacts
\cite{kwon2022catnet}. Compression physics is signal when modelled deliberately and
shortcut when merely correlated with the label. Such localisers train on mask datasets
\cite{dong2013casia,novozamsky2020imd2020,jia2023autosplice,zhang2023magicbrush}.
Closest to our setting, \cite{tantaru2024weakloc} train on image-level labels only and
evaluate the resulting maps against local diffusion edits. They conclude that
weakly-supervised localisation is attainable but degrades under generator and dataset
shift. SIDA \cite{huang2025sida} pairs detection with mask prediction on a 300k-image
social-media corpus, and adds a separately tuned explanation stage. So does
\cite{zhou2025aigiholmes}, with a multimodal language model. We train no localiser and
propose no new one. Our question is not how well a trained localiser segments an edit.
It is which off-the-shelf attribution method to place over a frozen detector.

\section{Dataset Details}
\label{sec:supp-data}

\paragraph{Real class.}
The real class comes from public photo reviews of a marketplace.
Only publicly visible content enters the corpus.
We store the images at full size, and they arrive already in WebP, because the service
that hosted them re-encoded them into that format.

Four filters run over the collected pool.
A face gate \emph{discards} frames with faces rather than blurring them (InsightFace
SCRFD \texttt{buffalo\_l}, multi-scale 640/320, threshold 0.5).
We discard rather than blur because blur would create an artefact correlated with the
class.
We also exclude frames already marked as blurred.
We deduplicate by product and by photo, which
dropped 249 duplicates at assembly.
A CLIP QC gate rejects screenshots, interfaces and text documents
(\texttt{clip-vit-base-patch32}, thresholds 0.65 / 0.75 at a maximum product score of
0.35).

From a pool of 200{,}000 collected frames we requested 170{,}000.
After deduplication 169{,}751 remained: 16{,}776 mandatory generation references and
152{,}975 additional frames for diversity.

\paragraph{Synthetic class.}
We produce synthetic images with our batch generator through OpenRouter.
The scheme is ``one reference $\rightarrow$ one model $\rightarrow$ one composite
prompt $\rightarrow$ one image''.
We draw the model deterministically from the pair identifier with weights
$w \propto 10^{\mathrm{elo}/400} / \mathrm{cost}^{0.684}$.
We calibrated those weights to a budget of about \$0.035 per image at measured prices.
The actual figure is $\approx$\,\$0.038.

We assemble prompts from independent groups: environment, lighting, presence of hands,
watermark, one of eight viewpoints, product defect, glare, pattern.
The synthetic class therefore imitates a review photo rather than a studio card.
We inject a defect (crack, stain, wear) into 20\% of object and 40\% of apparel
generations.
We normalise references to PNG 750$\times$1000 before submission.
We convert outputs to JPEG q85 and centre-crop them to exactly 3:4.

Class labels come from provenance rather than from human annotation.
Every generation carries a record of model, reference, category and all prompt parts.
Humans do poorly on this task: in an informal quiz we ran among ourselves, without a
controlled protocol, they separated fakes with $\approx$60--70\% accuracy.
Each synthetic image also inherits the product group of its reference, which makes a
leakage-free split possible.

The short names in Tab.~\ref{tab:supp-generators} abbreviate OpenRouter slugs.
Three of them are \texttt{gpt\_image\_2} = \texttt{openai/gpt-image-2}, \texttt{flux} =
\texttt{black-forest-labs/flux.2-klein-4b} and \texttt{flux\_pro} =
\texttt{black-forest-labs/flux.2-pro}. Two more are \texttt{nano\_banana\_lite} =
\texttt{google/gemini-3.1-flash-lite-image} and \texttt{nano\_banana\_2} =
\texttt{google/gemini-3.1-flash-image}.
The rest are \texttt{nano\_banana\_pro} = \texttt{google/gemini-3-pro-image},
\texttt{riverflow} = \texttt{sourceful/riverflow-v2.5-fast} and \texttt{seedream} =
\texttt{bytedance-seed/seedream-4.5}. The last two are \texttt{mai\_image} =
\texttt{microsoft/mai-image-2.5} and \texttt{grok\_imagine} =
\texttt{x-ai/grok-imagine-image-quality}.

\begin{table}[t]
\centering
\small
\setlength{\tabcolsep}{4pt}
\caption{Generator mix of the synthetic class. Actual shares deviate from the plan by at most 0.6 percentage points.}
\label{tab:supp-generators}
\begin{tabular}{lrrr}
\toprule
Generator & Images & Actual & Plan \\
\midrule
\texttt{gpt\_image\_2} & 2{,}886 & 17.2\% & 17.11\% \\
\texttt{flux} & 2{,}705 & 16.1\% & 16.01\% \\
\texttt{nano\_banana\_lite} & 1{,}944 & 11.6\% & 11.27\% \\
\texttt{riverflow} & 1{,}852 & 11.0\% & 10.98\% \\
\texttt{seedream} & 1{,}596 & 9.5\% & 10.06\% \\
\texttt{nano\_banana\_2} & 1{,}549 & 9.2\% & 9.02\% \\
\texttt{flux\_pro} & 1{,}518 & 9.0\% & 9.17\% \\
\texttt{mai\_image} & 1{,}243 & 7.4\% & 7.73\% \\
\texttt{grok\_imagine} & 1{,}195 & 7.1\% & 6.94\% \\
\texttt{nano\_banana\_pro} & 288 & 1.7\% & 1.71\% \\
\midrule
Total & 16{,}776 & 100\% & \\
\bottomrule
\end{tabular}
\end{table}

By product type the synthetic class is 15{,}520 object generations (92.5\%) and
1{,}256 apparel generations (7.5\%).
It covers 1{,}927 unique marketplace categories.

\paragraph{Splits and dataset versions.}
The split is two-level and cuts by product, never by image.
The collector first splits product groups 80/20,
stratified by the presence of synthetic images in the group.
A product goes entirely to one side.
The training script then takes the first fold of
\texttt{StratifiedGroupKFold(n\_splits=5, shuffle, random\_state=seed)}.
That gives an internal 80/20 train/validation split of 119{,}450 and 29{,}765
images, as recorded in the bundle's own metadata.
We never split at random.

We enforce the absence of leakage in code.
The splitter computes the group intersection and raises on any non-zero overlap.
For the external held-out test it separately counts the overlapping groups and rows
before it may mark the split product-disjoint.
It also checks that a reference lives in the same dataset as its generations.
The actual group intersection between train and test is zero, and the synthetic shares
are close: 8.9\% vs.\ 9.2\%.

We keep the 1:10.1 class imbalance near the product distribution.
PR-AUC is therefore the primary metric rather than accuracy: at a 9\% synthetic share
the trivial answer ``everything is real'' already gives 91\% accuracy.
We select the decision threshold on validation only and transfer it to the test as is.
Metrics with a threshold tuned on the test itself we do not publish.
R@P95 is not one of those, as Sec.~\ref{sec:data} states.
Rates that do depend on the selected threshold are always labelled ``at the validation
threshold''.

Both versions live in the content-addressed registry and are immutable.
A dataset id is the SHA-256 of its content: \texttt{dataset-v2-train}
\texttt{11876f70\ldots aaf8} and \texttt{dataset-v2-test} \texttt{8434ebf1\ldots ff6b}.
The job mounts a version read-only and verifies every file byte-wise before start.
We keep the early build \texttt{dataset-v2-17k} (186{,}527 images, no held-out) in the
registry as a historical artefact and retire it from this work.

\begin{table}[t]
\centering
\small
\setlength{\tabcolsep}{4pt}
\caption{Dataset versions (\texttt{dataset-v2-train}, \texttt{dataset-v2-test}). Groups are marketplace products; group intersection between train and test is zero. On-disk sizes are 17.0\,GB (train), 4.3\,GB (test) and 21.3\,GB (total).}
\label{tab:supp-splits}
\begin{tabular}{lrrrr}
\toprule
Dataset & Images & Real & Fake & Groups \\
\midrule
Train & 149{,}215 & 135{,}865 & 13{,}350 & 11{,}105 \\
Test & 37{,}312 & 33{,}886 & 3{,}426 & 2{,}885 \\
\midrule
Total & 186{,}527 & 169{,}751 & 16{,}776 & 13{,}990 \\
\bottomrule
\end{tabular}
\end{table}

Real/fake shares are 91.1\%/8.9\% in train and 90.8\%/9.2\% in test.

\paragraph{Localisation set.}
\texttt{dataset-v2-test-xai-webp} holds 368 triples of background, edit and mask,
about 209\,MB.
Of these, 183 insert an object into a scene (gpt-image-2, ``reference + background'').
The other 185 use mask-guided inpainting: 135 from flux and 50 from bria via fal.ai.

We link triples by \texttt{base\_id} and store masks losslessly, binarised at
$\ge$128.
The size reference inside a triple is the generation.
We resample masks NEAREST and backgrounds LANCZOS, and auto-transpose rotated frames.
Before publication we check the number of masks against the number of synthetic frames.
We register masks as metadata and verify them byte-wise.

The design specification asked for three things.
Originals had to be drawn only from held-out products.
Composition had to be strict: alpha blending in linear RGB, a lossless PNG master, a
byte-wise equality check outside the edit region, and only then lossy variants.
Each triple had to carry three masks, the requested \texttt{M\_requested}, the actually
applied \texttt{A\_applied} and the diagnostic \texttt{M\_realized}.

The assembled v1 meets none of the three, because we built it by merging six
heterogeneous external folders.
First, the provenance of originals is unconfirmed: nobody checked their correspondence to
held-out products.
We therefore keep the set under a separate alias and do not merge it into the main
test.
Second, strict composition is not guaranteed.
Frames are stored in lossy WebP q92, and the background outside the mask need not be
bit-identical to the original.
Third, only the reference mask is filled.
The fields \texttt{edit\_scope}, \texttt{task}, \texttt{edit\_type} and
\texttt{realized\_mask} are absent.
We therefore infer the edit type from the generator family (``gpt $\rightarrow$
insertion, fal $\rightarrow$ inpainting'') and do not compute \texttt{rma\_realized} on
this version.

Later measurements surfaced one further defect.
At least one pair (\texttt{xai\_0163}) carries inverted detector verdicts: the ``real''
side receives generation probability 0.9999 and the inpainted side 0.03.
That points to a possible provenance error on the real side.

Every mask here is the reference intervention mask of Sec.~\ref{sec:protocol}, not ``true AI
pixels''.

\paragraph{Storage format as a leakage channel.}
Real frames arrive already in WebP and synthetic ones in JPEG from the
generator.
A naive assembly therefore puts an almost perfect separating feature into the data.
This produced a sequence of data versions.

\textbf{v2} recompresses synthetic images to WebP
(\texttt{dataset-v2-\{train,test\}-webp}) and copies real frames byte-wise, which
weakens the shortcut without removing it.
\textbf{v2-webp2} applies \emph{the same final encode to both classes}: RGB, then
LANCZOS to 1200\,px on the long side, then WebP q80 at method 4.
It carries a recompression provenance block per manifest row and is published as a new
immutable dataset.
The \textbf{native control} keeps synthetic images in their original JPEG and real
frames as they are.
That is the distribution of files as users upload them, and we declare it the decisive
measurement.

We also implemented a symmetric JPEG control (\texttt{jpeg2}), which needs a guard of
its own.
JPEG uses an 8$\times$8 DCT grid anchored to the origin.
After the WebP stage real frames lose their original grid, while synthetic images,
originally JPEG, keep it.
A naive symmetric JPEG would therefore create a \emph{new} potential shortcut,
``aligned double JPEG = fake''.
We mitigate it with a random 0--7 pixel shift before encoding, which desynchronises the
grids of both classes.
This control exists as code only.
We have no run results and report no numbers for it.

One rule stands behind all of these versions.
We apply any transform that changes the codec (JPEG, resize, blur) \emph{symmetrically to both
classes}.

\paragraph{Known dataset biases.}
Five biases survive in this corpus and we did not remove them.
\textbf{Category skew}: nine of the ten most frequent synthetic categories are sports
nutrition, about 27\% of all generations.
The risk is learning ``protein jar = fake'' instead of generative traces, and we did
not run a per-category metric slice.
\textbf{Thin generator tail}: \texttt{nano\_banana\_pro} has only 288 images (1.7\%),
of which about 58 land in the held-out test.
That is too few for stable per-generator analysis.

\textbf{Two generation batches}: we collected the synthetic class in two runs of
11{,}815 and 4{,}961 images.
For the second batch we reconstructed some references from input pairs and marked them
with a separate flag.
We did not measure whether the batches are distinguishable.
\textbf{Balance shift between versions}: moving to \texttt{dataset-v2} changed the
balance from about 48:1 to 10.1:1.
That required retuning the positive class weight (6.0 $\rightarrow$ 3.2) and makes
numbers from before and after the transition incomparable.
\textbf{Single source}: all real frames come from review photos of one marketplace and
carry the processing history of the service that hosted them.
We did not test transfer to other sources.

\begin{table}[t]
\centering
\footnotesize
\caption{Per-generator PR-AUC on the held-out test, from the two later full-feature runs (2026-07-20 and 2026-07-21). Both train on the native training set, so the format channel inflates the absolute level and it
is not comparable with the main results. The spread across generators is the informative part.}
\label{tab:supp-pergen}
\begin{tabular}{lrrr}
\toprule
Generator & images & run A & run B \\
\midrule
\texttt{flux} & 2{,}705 & 0.994 & 0.991 \\
\texttt{seedream} & 1{,}596 & 0.988 & 0.985 \\
\texttt{grok\_imagine} & 1{,}195 & 0.977 & 0.975 \\
\texttt{gpt\_image\_2} & 2{,}886 & 0.976 & 0.959 \\
\texttt{nano\_banana\_lite} & 1{,}944 & 0.971 & 0.960 \\
\texttt{nano\_banana\_2} & 1{,}549 & 0.967 & 0.962 \\
\texttt{mai\_image} & 1{,}243 & 0.964 & 0.948 \\
\texttt{riverflow} & 1{,}852 & 0.955 & 0.946 \\
\texttt{flux\_pro} & 1{,}518 & 0.910 & 0.882 \\
\texttt{nano\_banana\_pro} & 288 & 0.859 & 0.837 \\
\midrule
overall & 16{,}776 & 0.993 & 0.989 \\
ECE (15 bins) & --- & 0.0050 & 0.0057 \\
\bottomrule
\end{tabular}
\end{table}

The two runs agree on the ordering.
The two weakest generators are the two with the fewest training images:
\texttt{flux\_pro} shares its family with \texttt{flux}, and
\texttt{nano\_banana\_pro} contributes 1.7\% of the synthetic class.
On the symmetrically re-encoded test the same runs give PR-AUC 0.969 and 0.966, with
expected calibration error (ECE) 0.0108 and 0.0114.
ECE is the average gap between stated confidence and observed accuracy.
Calibration therefore degrades by roughly a factor of two when the evaluation axis
changes.
A threshold tuned on one axis does not transfer unchanged to another.

\begin{figure}[t]
  \centering
  \includegraphics[width=\columnwidth]{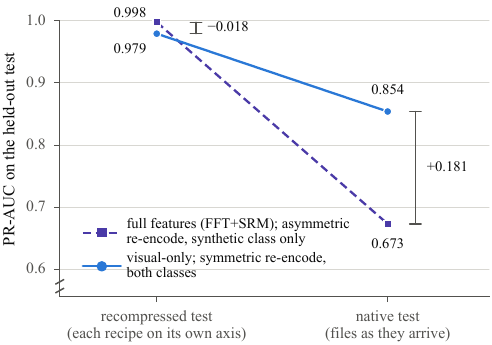}
  \caption{What the two remedies did to PR-AUC. The champion keeps the forensic features and saw data where only the synthetic class was re-encoded. The second-fix recipe uses visual-only features and data where both classes pass the same final encode. Each line's recompressed point is its own axis, so the 0.018 gap is a change of operating point rather than a controlled delta. The native test is the same for both, and it is the distribution of files as users upload them.}
  \label{fig:tradeoff}
\end{figure}

\begin{figure}[t]
  \centering
  \includegraphics[width=\columnwidth]{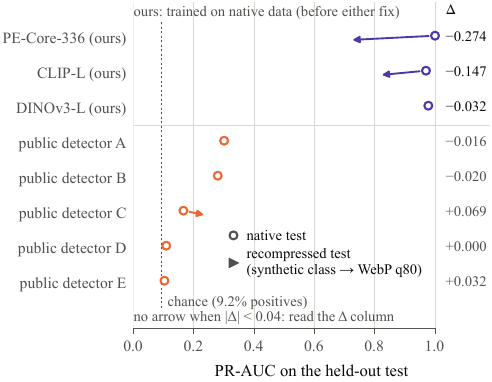}
  \caption{How much of each score was compression format. Hollow circle: PR-AUC on the native test. Arrow head: the same test with the synthetic class re-encoded to WebP q80, the real class's format. Our three backbones lose up to 0.274 while five public detectors move by at most 0.07 in either direction.}
  \label{fig:profile}
\end{figure}

\section{The Shortcut Runs in Detail}
\label{sec:supp-shortcut}

This section holds the run-level detail behind Sec.~\ref{sec:shortcut}, and
Fig.~\ref{fig:tradeoff} places the two remedies side by side on both axes.

\paragraph{The alarm.}
The stored PR-AUC of the model trained on the corpus as collected rounds to 1.0000 at
four places, and we print the conservative 0.9999 throughout.
On the WebP axis DINOv3-L loses only $-0.032$ PR-AUC and still reaches R@P95 0.816,
against $-0.274$ for PE-Core-336 and $-0.147$ for CLIP-L.
We list the five public detectors with their PR-AUC on the native and the WebP test.
\texttt{haywoodsloan/ai-image-detector-deploy} (SwinV2) reads 0.300 / 0.284, and
\texttt{Ateeqq/ai-vs-human-image-detector} (SigLIP) 0.279 / 0.259.
\texttt{Smogy/SMOGY-Ai-images-detector} (Swin) reads 0.165 / 0.234.
\texttt{NYUAD-ComNets/\allowbreak NYUAD\_AI-generated\_images\_detector} (ViT) reads 0.108 /
0.108, and \texttt{Organika/sdxl-detector} (Swin) 0.102 / 0.134.
We loaded each through \texttt{transformers} from its Hugging Face default branch during
benchmark run \texttt{7cd86b6e} on 2026-07-17, with its own preprocessing and with no revision
pinned. The missing revision pin is a reproducibility gap of ours.
They move by at most 0.07 in either direction under the same transformation, so WebP
re-encoding does not break detectors as such.

\paragraph{The audit behind the first fix.}
Retrained on \texttt{dataset-v2}, PE-Core-336 reaches PR-AUC 0.9976 on the recompressed
axis at ROC-AUC 0.9998 and R@P95 0.993, where the best of the five public detectors
reaches 0.284.
A hand check on a few dozen files then located the failure.
The model was right on real frames as their host had re-encoded them, flagging 0 of 15.
It was right on fakes recompressed by our own re-encoder ($P>0.99$).
On the same fakes in their native \texttt{.jpg} the median $P(\text{fake})$ was 0.008,
and 5 of 20 were caught.
Inpainted edits scored $P=0.0$.
Every file we pushed through our own re-encoder was labelled fake, real photographs
included.
We then measured the same effect on the full native test set, at median $P(\text{fake})$
0.0004 and 24.7\% caught at the validation threshold.
We keep the hand check and the measurement apart on purpose.

\paragraph{Bookkeeping for the factorial.}
A first attempt at the asymmetric/visual-only cell does not enter the design, because it had
silently fallen back to the spec's default backbone. Defaults are a provenance channel too.
We retrained ten of the twelve runs independently of our cluster, on rented single-GPU
machines, from the published datasets and lock files.
The cell containing the second-fix model reproduces its reported numbers exactly at seed
42 (0.8540 native, 0.9792 recompressed).
The old champion's number is reproduced to within 0.007.
One reproducibility seam belongs here.
We trained the final model outside the scheduled loop, on a rented bare-metal GPU. Its
environment therefore sits under a different lock file than the one its bundle carries
(\texttt{transformers} 4.48.3 against 4.57.6).

\paragraph{The feature-channel probe.}
Sec.~\ref{sec:shortcut} quotes three numbers from this probe, and
Tab.~\ref{tab:supp-channels} holds the rest.
We fit one head architecture on disjoint channels of a single PE-Core-336 pass over
42{,}991 images and read it on a product-disjoint test whose share of synthetics is
0.254.
Three corpora enter. The asymmetric one re-encodes synthetics only, and the symmetric one puts
both classes through the same final encode. The crossed one assigns each file's final container
by a hash of its name, independent of the label.
Two readings follow.
The forensic channel decays across the three, 0.9252 to 0.7145 to 0.5927 on validation.
Symmetric encoding weakens it, and randomising the container removes about half of what
survives.
The full model barely moves over the same three, 1.0000 to 0.9935 to 0.9901, and the
forensic block buys it $+0.0009$ on the symmetric corpus.
A separating compression channel therefore outlives the fix, and the model we selected
does not depend on it.

\begin{table}[t]
\centering
\footnotesize
\setlength{\tabcolsep}{4pt}
\caption{The feature-channel probe. One head architecture ($d\to256\to64\to1$) fitted per
row on disjoint channels of a single PE-Core-336 pass over 42{,}991 training images, read
on a product-disjoint test. \emph{visual} is the pooled pre-logit embedding,
\emph{forensic} the 44 hand-designed features (32 FFT radial bins and 12 SRM statistics),
\emph{both} their concatenation. PR-AUC on the validation split and on the native test.
Chance is the synthetic base rate, 0.254. \emph{asym.}: only synthetics re-encoded, the
first fix. \emph{sym.}: one identical final encode for both classes, the second fix.
\emph{crossed}: the final container assigned per file by a hash of its name, independent
of the label.}
\label{tab:supp-channels}
\begin{tabular}{llcc}
\toprule
Corpus & Channel & validation & native \\
\midrule
asym.\  & visual   & 1.0000 & 1.0000 \\
        & forensic & 0.9252 & 0.9198 \\
        & both     & 1.0000 & 1.0000 \\
\midrule
sym.\   & visual   & 0.9926 & 0.9893 \\
        & forensic & \textbf{0.7145} & 0.6890 \\
        & both     & 0.9935 & 0.9909 \\
\midrule
crossed & visual   & 0.9893 & 0.9848 \\
        & forensic & 0.5927 & 0.5536 \\
        & both     & 0.9901 & 0.9865 \\
\bottomrule
\end{tabular}
\end{table}

\paragraph{Invariance to encoding history.}
The probe above asks what the container still carries, and this run asks whether a model
follows it.
We scored 1{,}500 real and 1{,}500 synthetic native test files under eight encoding
histories in turn, and read two bundles on each (Tab.~\ref{tab:supp-history}).
\texttt{png} is pixel-lossless, so it has to reproduce the native row exactly, and it
does; that equality is the run's own self-check.
Throughout this appendix we write \emph{sealed} for the first-fix bundle of
Sec.~\ref{sec:shortcut}.
Across the eight histories the sealed bundle spans 0.1484 PR-AUC and the selected one
0.0507.
Per image the difference is starker, but neither model is invariant and the median alone
overstates the selected one.
The median sealed image moves \textbf{0.9988} of the probability range as the container
changes, at an interquartile range of [0.014, 1.000] and a mean of 0.648. The selected median
moves \textbf{0.0001}, but at an interquartile range of [0.000, 0.990] and a mean of 0.328.
59.0\% of images flip predicted class under some history for the sealed bundle against
33.0\% for the selected one, and restricted to synthetics those rates are 97.5\% and
59.1\%.
We therefore report median, spread and flip rate together and do not describe the
selected model as invariant to encoding history.
This run is 1{,}500 real and 1{,}500 synthetic, a 50\% base rate, so its PR-AUC is not on
the scale of the 8.9\%-prevalence numbers elsewhere in this paper.
The crossed test then gives each file a history drawn from a hash of its name, so the container
carries no label information at all. The sealed bundle falls to 0.8459 while the selected
one reads 0.9769, level with its own native axis.

\begin{table}[t]
\centering
\footnotesize
\setlength{\tabcolsep}{4pt}
\caption{PR-AUC by encoding history. 1{,}500 real and 1{,}500 synthetic native test files,
each scored under every history in turn. \emph{Sealed} is the champion of
Sec.~\ref{sec:shortcut}, asymmetric \texttt{webp} data with the forensic features on, and the
bundle behind Sec.~\ref{sec:supp-firstfix}. \emph{Selected} is symmetric \texttt{webp2} data
with the forensic features on, the factorial cell we chose. \texttt{png} is
pixel-lossless and reproduces the native row exactly, which is the run's self-check.
\emph{Spread} is the largest minus the smallest of the eight histories. \emph{Median
range} is the per-image range of $P(\text{fake})$ across them. \emph{Crossed} assigns each
file its history by a hash of its name, independent of the label.}
\label{tab:supp-history}
\begin{tabular}{lcc}
\toprule
Encoding history & sealed & selected \\
\midrule
native (as uploaded) & 0.9116 & 0.9752 \\
\texttt{webp80} & 0.9859 & 0.9973 \\
\texttt{jpeg85} & 0.8391 & 0.9467 \\
\texttt{jpeg85}$\to$\texttt{webp80} & 0.9798 & 0.9931 \\
\texttt{webp80}$\to$\texttt{webp80} & 0.9854 & 0.9971 \\
\texttt{jpeg50} & 0.8675 & 0.9803 \\
\texttt{resize1200}$\to$\texttt{webp80} & 0.9876 & 0.9970 \\
\texttt{png} (lossless) & 0.9116 & 0.9752 \\
\midrule
Spread across histories & 0.1484 & \textbf{0.0507} \\
Median per-image range & 0.9988 & \textbf{0.0001} \\
Images flipping class & 59.0\% & \textbf{33.0\%} \\
\quad synthetics only & 97.5\% & \textbf{59.1\%} \\
Crossed histories, PR-AUC & 0.8459 & \textbf{0.9769} \\
\bottomrule
\end{tabular}
\end{table}

\section{Detector: Calibration, Encoder and Unseen Generators}
\label{sec:supp-detector}

\paragraph{Calibration and per-generator spread.}
A decision threshold needs a calibrated score.
We therefore Platt-scale the head on the validation split \cite{guo2017calibration} and
read every number here off the calibrated bundle.
In two later runs we measured the calibration error directly. ECE over
15 bins is 0.0057 on the native test and 0.0114 on the symmetrically re-encoded test.
Those runs train on the native training set, \ie{} on the axis where the format channel
is still present.
Their PR-AUC (0.989--0.993 native) is therefore not comparable with Tab.~\ref{tab:detector},
and we do not quote it as a result.
What we read from them is the calibration error and the per-generator spread.

Per-generator PR-AUC on the held-out test ranges from 0.837 for
\texttt{nano\_banana\_pro} to 0.994 for \texttt{flux} (Tab.~\ref{tab:supp-pergen}).
\texttt{nano\_banana\_pro} is the thinnest class, 288 images in total.
The weakest generators are the ones with the fewest training examples rather than the
newest ones.
The factorial cell that keeps the forensic features is the one scheduled run with
per-generator metrics on the native axis.
On it PR-AUC ranges from 0.954 for \texttt{riverflow} down to 0.366 for
\texttt{mai\_image}, at ECE 0.0080 recompressed and 0.0287 native.
The two later runs and that cell are not comparable cell by cell, because the later runs
train on the native training set and the factorial cell does not.

\paragraph{Descriptor and head in full.}
The FFT block takes $\log(1+|\mathrm{FFT2}|)$ of the grey central $224\times224$ crop at
native resolution, zeroes the DC term and averages the spectrum into 32 radial bins.
We never upscale, since interpolation would synthesise the very statistics we measure.
The SRM block applies three fixed $5\times5$ high-pass kernels and takes four statistics
per residual, so the two blocks give $32{+}12=44$ forensic features.
The head is an MLP ($d\to256\to64\to1$, dropout 0.2) with $\mathrm{pos\_weight}=3.2$,
stopped early on validation PR-AUC at seed 42.

\paragraph{Is the frozen encoder doing the work?}
We ran no end-to-end fine-tuning of the selected architecture, so we do not bound the
alternative by measurement.
The choice rests on three things instead.
13{,}350 synthetic training images across ten generators do not constrain a large ViT,
so a fine-tuned encoder would have room to fit those ten generators rather than
synthesis.
Frozen features over a pretrained encoder are the stronger reported baseline for this
task \cite{ojha2023universal,cozzolino2024raisingbar}.
And freezing is what makes the ablations in this paper readable.
With the representation held fixed, only the head and the feature blocks can move a
number.

\paragraph{Unseen generators.}
Leave-one-generator-out excludes a generator from training entirely and uses it as a
blind test.
On the model \emph{before} the second fix, six of ten single-generator folds gave
PR-AUC 0.9904--0.9998.
Those folds ran with $\mathrm{pos\_weight} = 6.0$, the value from before the class
balance changed, rather than the 3.2 of Sec.~\ref{sec:detector}.
The two LOGO generations are therefore less comparable than the shared name suggests.
We never believed that number, because a format trace transfers between generators
trivially and sibling models leak into each other's folds.
We therefore re-ran LOGO with whole \emph{families} held out at once, so a held-out
generator cannot be seen through its sibling (Tab.~\ref{tab:supp-logo}).
All three folds train on \texttt{dataset-v2-train-webp2} and evaluate on
\texttt{dataset-v2-test-webp2}, the recompressed axis rather than the native one.

All three folds now run on the selected recipe.
Held-out families give 0.979, 0.921 and 0.919 PR-AUC, at R@P95 0.912, 0.727 and 0.736.
Transfer to an unseen family is therefore real, and it costs between two and eight points
against in-distribution validation.
Two of those folds first read 0.924 and 0.850, while a configuration default had selected a
CLIP-L backbone. We therefore read our earlier claim that the nano-banana family is the hardest
as a property of the default rather than of the data.

Three caveats belong with the table.
The leakage guard is aggressive: a held-out fake whose product group appears in
training is dropped.
That removes roughly three quarters of the holdout fakes and leaves 629--815 per fold.
These numbers are therefore computed on a small, adversarially selected slice.
We ran one fold per family, one seed.

\begin{table}[t]
\centering
\footnotesize
\setlength{\tabcolsep}{4pt}
\caption{Family-level leave-one-generator-out on the selected recipe: PE-Core with the forensic features
on. We train on \texttt{dataset-v2-train-webp2} and evaluate on \texttt{dataset-v2-test-webp2},
the recompressed axis. We exclude whole generator families from training, then drop held-out fakes whose product group
appears in training. The usable count is therefore far below the family's holdout size: 815 of
3{,}380 for \texttt{flux}, 629 of 2{,}288 for \texttt{gpt\_image\_2}, 707 of 3{,}003 for
\texttt{nano\_banana}. The last two rows are the same folds as first published, when a configuration default had
selected a CLIP-L backbone. We keep them to show what such a default costs. Read together, they retire our earlier reading that the nano-banana family is the hardest. On
one architecture it looks worst by seven points; on the other the three families sit within
six.}
\label{tab:supp-logo}
\begin{tabular}{llcccc}
\toprule
Held-out family & backbone & PR-AUC & R@P95 & ECE & fakes \\
\midrule
\multicolumn{6}{l}{\emph{Selected recipe: PE-Core, forensic features on}} \\
\texttt{flux}$+$\texttt{flux\_pro} & PE-Core & \textbf{0.979} & \textbf{0.912} & 0.013 & 815 \\
\texttt{gpt\_image\_2} & PE-Core & \textbf{0.921} & \textbf{0.727} & 0.037 & 629 \\
\texttt{nano\_banana\_*} & PE-Core & \textbf{0.919} & \textbf{0.736} & 0.028 & 707 \\
\midrule
\multicolumn{6}{l}{\emph{As first published, on a backbone taken from a default}} \\
\texttt{flux}$+$\texttt{flux\_pro} & CLIP-L & 0.924 & 0.749 & 0.019 & 815 \\
\texttt{nano\_banana\_*} & CLIP-L & 0.850 & 0.598 & 0.034 & 707 \\
\bottomrule
\end{tabular}
\end{table}

\begin{table}[t]
\centering
\small
\caption{Second fix. The champion (full forensic features, asymmetric \texttt{webp}
data) against the visual-only model trained on symmetrically encoded \texttt{webp2}
data. The two recompressed rows are each recipe's own axis, so they are different test
sets and not comparable. The native rows are the same test for both models, and that is
the distribution of files as users upload them.}
\label{tab:detector}
\begin{tabular}{lcc}
\toprule
Axis & Champion (full) & Visual-only \\
\midrule
Recompressed, PR-AUC            & 0.9976 & 0.9792 \\
Recompressed, R@P95             & 0.993  & 0.916 \\
Native, PR-AUC                  & 0.6732 & \textbf{0.8540} \\
Native fakes, median $P$(fake)  & 0.0004 & \textbf{0.9999} \\
Native fakes, caught @ val thr. & 24.7\% & \textbf{68.6\%} \\
Native, R@P95                   & 0.236  & \textbf{0.609} \\
False positives on real frames  & \textbf{0.15\%} & 0.76\% \\
\bottomrule
\end{tabular}
\end{table}

\section{Evaluation Protocol in Full}
\label{sec:supp-protocol}

We treat an explanation as a measured output and evaluate it along four dimensions.
All methods and all metrics use the \emph{same patch geometry and the same replacement
strategy}, defined in a shared perturbation module.
The grid always sits inside the crop the model actually sees.
We aggregate \emph{macro over \texttt{base\_id}}, taking the mean of per-group means
rather than a flat mean over rows.
Confidence intervals come from a two-stage cluster bootstrap over groups with 2000
resamples, read at the 2.5/97.5 percentiles.

\paragraph{(1) Faithfulness.}
This dimension collects established quantities from the literature.
AOPC, deletion AUC (lower is better) and insertion AUC (higher is better) follow
\cite{petsiuk2018rise}.
Comprehensiveness and sufficiency at $q = 0.2$ follow \cite{deyoung2020eraser}.
We compute comprehensiveness on a separate perturbation rather than reading it off the
deletion curve.
Infidelity \cite{yeh2019infidelity} uses a single scale coefficient fitted by least
squares, because attribution units are not comparable across methods.
SaCo \cite{wu2024saco} is the agreement between the importance ordering and the actual
influence, in $[-1, 1]$.

We build the target normalisation once, on the pair of original and fully perturbed
baseline.
On degenerate contrast we flag the example \texttt{insufficient\_target\_contrast}
instead of inflating the aggregates.
We publish not the absolute value but the \emph{per-image paired advantage over a
control}, with a bootstrap interval.
A method counts as working only if the lower bound of its advantage over
\texttt{random} exceeds zero.

\paragraph{(2) Sanity.}
We compare each map with the map of the same method computed on a head retrained on
shuffled labels \cite{adebayo2018sanity}.
High similarity, by Spearman or by IoU at top-$q$, means the method does not explain
the trained model.
The limits of this check are explicit.
We train the shadow head on features of the evaluation sample itself, and gradient methods do
not pass through it.
An empty value must therefore not be read as ``check passed''.
Cascading randomisation of encoder layers \cite{adebayo2018sanity} is not implemented.

\paragraph{(3) Stability and sensitivity.}
Stability is the agreement of the map under a weak input transform.
The main transform is JPEG q70; under a flip we mirror the map back before comparison.
We report Spearman $\rho$, IoU at top-$q$, relative mass change and the verdict-flip
rate.
Sensitivity is sensitivity-max, in the spirit of \cite{yeh2019infidelity}. We take the maximum
over samples of $1 - \rho$ between the map on the original and on a slightly noised input
($\sigma = 0.03$).
In the repository configuration both blocks run on a subsample of 10 images and exclude
\texttt{occlusion}/\texttt{rise} for cost.
The runs behind this paper use those parameters as they stand.
We therefore measure stability on 10 images under one transform, JPEG q70, and neither
perturbation method returns a stability value at all.

\paragraph{(4) Localisation metrics.}
We compute the following against the reference intervention mask.
\textbf{Pixel AP} is pixel-level average precision after
bilinear upsampling of the map.
We rank it by the \emph{signed} value, so negative evidence sinks to the bottom.
With no positives it returns NaN rather than a free 1.0.
\textbf{Patch AP} is the same quantity on the native grid of the method, where a cell
counts as positive at mask coverage $\ge$0.5.
\textbf{RMA} and the \textbf{pointing game} \cite{zhang2018pointinggame} are as defined in
Sec.~\ref{sec:protocol}.
\textbf{IoU and Dice} use a \emph{frozen} threshold top-$q = 0.2$, never tuned per
image.
Finally, we compute the relevance share in the core, ring and outside zones.
We build the ring by dilation and erosion with a radius of 1\% of $\min(H, W)$, and the
three zones partition the frame completely.
A map with no positive energy is flagged and receives no convenient score.

\paragraph{Two tables.}
Every method receives \emph{two} tables.
The \textbf{conditional} table covers valid maps only, and we always report it with
\texttt{n\_valid / N\_total}.
The \textbf{end-to-end} table gives a pre-registered zero to three cases: an invalid
map, a map with no positive energy, \emph{and a detector miss}.
A detector miss is a raw logit not above zero on a synthetic frame.
The second table closes a known loophole: a method cannot look strong by producing maps
only for its easiest true positives.
For our system the gap between the two tables is large, and that gap is itself a
result.
The first-fix bundle recognises 4 of 100 inpaintings on its edits axis and 20 of 368 on
the localisation set (Sec.~\ref{sec:supp-region}).
That gap is largely the bundle: the detector we selected recognises 142 of the same 368
(Tab.~\ref{tab:supp-served}) and 75 of the 200 the unified run samples.
The visual-only model of the second fix recognised 25 of 100 on a run of its own, which is
a different detector on a different run.

\paragraph{Slices.}
We additionally slice results by mask size, by edit type and by generator family.
The size buckets are tiny $\le$1\%, small $\le$5\%, medium $\le$15\%, large $\le$35\%
and xl.
The edit types are insertion and inpainting.
Masks smaller than the patch grid go into a \emph{separate stress slice} and are not
mixed into the main aggregates.

\paragraph{Acknowledged protocol weakness.}
The protocol has one weakness we state in advance.
The only replacement strategy is blur, and \texttt{occlusion} is itself built on blur.
This circularity gives the perturbation family an advantage.
We report a control run with mean replacement in Sec.~\ref{sec:supp-replacement}, and it
reverses the sign of the region family's advantage.
We compute descriptive shape statistics (Gini sparsity, normalised entropy, total variation)
per map.
They are not faithfulness metrics and enter no conclusion.

\begin{table}[t]
\centering
\small
\setlength{\tabcolsep}{4pt}
\caption{Explanation evaluation protocol hyperparameters.}
\label{tab:supp-protocol}
\begin{tabular}{ll}
\toprule
Parameter & Value \\
\midrule
Quantile schedule & \{0, 0.15, 0.3, 0.45, 0.6, 0.8, 1.0\} \\
Replacement & blur (default; mean run as a control) \\
\texttt{comp\_suff\_q} & 0.2 \\
Infidelity & 0.2 of cells, 8 samples \\
SaCo & 8 importance groups \\
Localisation & \texttt{top\_q} 0.2 (frozen); ring radius \\
 & 1\% of $\min(H,W)$; coverage 0.5 \\
Sanity / stability & 10 images each; $\sigma = 0.03$, \\
\ \ / sensitivity & 2 samples; transform JPEG q70 \\
Bootstrap & 2000 resamples, clustered \\
 & by \texttt{base\_id}, CI 2.5/97.5 \\
Controls & \texttt{random}, \texttt{center}, \texttt{edge} \\
 & (added automatically) \\
\bottomrule
\end{tabular}
\end{table}

\section{Method Registry}
\label{sec:supp-registry}

The registry holds 28 implementations.
Experimental methods passed only a smoke run on three images.
Beside the three mandatory controls, in the 18-method run we measure black-box perturbation
methods \cite{zeiler2014occlusion,petsiuk2018rise} and attention diagnostics
\cite{abnar2020rollout}. We also measure gradient and CAM methods
\cite{selvaraju2017gradcam,jiang2021layercam,chattopadhay2018gradcampp,sundararajan2017ig,smilkov2017smoothgrad}.
Our two pooling methods and our four region methods complete the set.
The attention diagnostics read no trained head, so they must survive head randomisation
unchanged.
We omit Score-CAM \cite{wang2020scorecam} and KernelSHAP/LIME
\cite{lundberg2017shap,ribeiro2016lime} on cost.
Recent ViT-specific work \cite{xie2023vitcx,bousselham2024legrad,wu2024tokentm} sits in the
default benchmark configuration, but we did not include it in the 18-method run we report here.

\paragraph{Our methods in detail.}
\texttt{pool\_decomposition} is a \emph{first-order} attribution.
The decomposition of the pooled vector is exact, while the attribution of the logit is
not, because a non-linear head sits after the pooling.
What is exact is the sum rule over tokens at first order, and our tests check it.
\texttt{pool\_occlusion} replays all token ablations in one batch through a small
pooling and head stack.
We validate each pooling method's replica against the detector's own embedding, and the
method returns \texttt{invalid} otherwise.
Both also localise modestly, because a single token's contribution is diluted by depth.
\texttt{object\_region} reads only the raw logit, so any head works.
\texttt{compact\_region} picks the smallest area budget in $\{0.05,\ldots,0.30\}$ that
reaches 80\% of the maximal effect.
SLIC \cite{achanta2012slic} is cheap and lets us control the number of units.
Of the three region methods, \texttt{object\_region} leads on relevance mass at the lowest cost
(Fig.~\ref{fig:region}). The ensemble leads on pixel AP at about twice the cost per frame, and the
trivial controls still lead on pointing.
The chronology of those choices is in Sec.~\ref{sec:ours}; the segment counts are 44 for the
\texttt{object\_region} variant and 48 for the ensemble we selected.
Tab.~\ref{tab:supp-served} scores both the raw grid and the display variant a viewer sees, on
both detectors.

\begin{table}[t]
\centering
\small
\setlength{\tabcolsep}{4pt}
\caption{Attribution method registry (28 implementations). ``Bench'' = member of the default benchmark configuration of 14; we measure a different set in the 18-method run of Tab.~\ref{tab:methods}, and another in the
15-method run of Tab.~\ref{tab:supp-firstfix}. ``ours'' = developed in this work; ``exp.'' = smoke run only.}
\label{tab:supp-methods}
\begin{tabular}{lll}
\toprule
Method & Type & Status \\
\midrule
\texttt{random} & control & bench \\
\texttt{center} & control & bench \\
\texttt{edge} & control & bench \\
\texttt{occlusion} & perturb. & bench \\
\texttt{rise} & perturb. & bench \\
\texttt{attention\_rollout} & attention & bench \\
\texttt{raw\_attention} & attention & bench \\
\texttt{chefer\_relevance} & attn+grad & off-bench \\
\texttt{vit\_gradcam} & gradient & bench \\
\texttt{layercam} & gradient & bench \\
\texttt{gradcam\_plusplus} & gradient & bench \\
\texttt{integrated\_gradients} & gradient & bench \\
\texttt{smoothgrad} & gradient & off-bench \\
\texttt{vit\_cx} & causal & bench \\
\texttt{legrad} & gradient & bench \\
\texttt{tokentm} & attention & bench \\
\texttt{pool\_decomposition} & ours & off-bench \\
\texttt{pool\_occlusion} & ours & off-bench \\
\texttt{object\_region} & ours & off-bench \\
\texttt{compact\_region} & ours & off-bench \\
\texttt{region\_ensemble} & ours & off-bench \\
\texttt{patch\_anomaly} & forensic & off-bench \\
\texttt{minimal\_flip} & exp. & exp. \\
\texttt{occlusion\_consensus} & exp. & exp. \\
\texttt{agreement\_map} & exp. & exp. \\
\texttt{boundary\_direction} & exp. & exp. \\
\texttt{causal\_refined} & exp. & exp. \\
\texttt{artifact\_fragility} & exp. & exp. \\
\bottomrule
\end{tabular}
\end{table}

Short descriptions follow, grouped by type.

\textbf{Controls.}
\texttt{random}: RNG map with a fixed seed.
\texttt{center}: Gaussian bump at the centre, \texttt{sigma\_frac} = 0.35.
\texttt{edge}: magnitude of the Sobel operator with area pooling.

\textbf{Perturbation and causal methods.}
\texttt{occlusion}: blur one patch at a time, relevance is the logit drop
\cite{zeiler2014occlusion}.
\texttt{rise}: random smooth masks, weighted sum of scores \cite{petsiuk2018rise}.
\texttt{vit\_cx}: clustering of patch embeddings (16) and perturbation of clusters
\cite{xie2023vitcx}.

\textbf{Attention methods.}
\texttt{attention\_rollout}: product of attention matrices across layers with the
residual connection \cite{abnar2020rollout}.
\texttt{raw\_attention}: CLS row of the last block without the layer chain.
\texttt{chefer\_relevance}: $\mathrm{ReLU}(\mathrm{grad} \times \mathrm{attention})$
with a rollout chain \cite{chefer2021transformer}.
\texttt{tokentm}: rollout weighted by the magnitude of token transformation
\cite{wu2024tokentm}.

\textbf{Gradient methods.}
\texttt{vit\_gradcam}: channel weights as the mean gradient over tokens
\cite{selvaraju2017gradcam}.
\texttt{layercam}: per-channel $\mathrm{ReLU}(\mathrm{grad})$ at every position
\cite{jiang2021layercam}.
\texttt{gradcam\_plusplus}: closed-form weight approximation from the first gradient
\cite{chattopadhay2018gradcampp}.
\texttt{integrated\_gradients}: path integral of the gradient, 32 steps, blurred
baseline \cite{sundararajan2017ig}.
\texttt{smoothgrad}: map averaged over noised copies of the input
\cite{smilkov2017smoothgrad}.
\texttt{legrad}: $\mathrm{ReLU}(\mathrm{grad})$ over post-softmax attention, mean of CLS
rows \cite{bousselham2024legrad}.

\textbf{Ours.}
\texttt{pool\_decomposition}: first-order decomposition of attention pooling into token
contributions, exact for the pooled vector and first-order for the logit.
\texttt{pool\_occlusion}: token ablation inside the pooling, all ablations in one batch.
\texttt{object\_region}: superpixel occlusion (SLIC \cite{achanta2012slic}), causal
score over regions.
\texttt{compact\_region}: extremal compact connected region under an area budget.
\texttt{region\_ensemble}: consensus of \texttt{object\_region},
\texttt{compact\_region} and \texttt{patch\_anomaly}, snapped to superpixels.

\textbf{Forensic and experimental.}
\texttt{patch\_anomaly}: deviation of a patch embedding from the mean, single pass.
\texttt{minimal\_flip}: binary search for the minimal set that flips the verdict.
\texttt{occlusion\_consensus}: agreement of two replacement strategies (blur, mean).
\texttt{agreement\_map}: geometric mean of several methods.
\texttt{boundary\_direction}: $\mathrm{grad} \times \mathrm{diff}$ at the sign-change
point along the baseline\,$\rightarrow$\,input path.
\texttt{causal\_refined}: shape from \texttt{legrad} inside the causal cluster weight of
\texttt{vit\_cx}.
\texttt{artifact\_fragility}: destroying the high-frequency fingerprint of a patch
instead of removing it.

\paragraph{Two methods that return nothing, and why.}
Two rows of Tab.~\ref{tab:supp-firstfix} carry dashes, for two different reasons, and both are
findings about the detector and the implementation rather than about the methods' merit.

\texttt{chefer\_relevance} returns 0 valid maps of 100, every one rejected as
\texttt{blank\_map}. It is not a crash and not a low score: the method emits an
identically zero relevance map. The core matrix chaining needs the gradient of the target with respect to each layer's
attention matrix. On this attention implementation every layer's gradient comes back
\texttt{None}. The tensors returned by \texttt{output\_attentions=True} carry \texttt{requires\_grad=True},
because differentiable operations produced them. They are not ancestors of the pooled output:
the model computes the returned weights on a path parallel to the one that produces the forward
hidden states. Verified for all 24 layers, independent of call order. The method therefore
reports a blank map instead of fabricating a plausible one from zero relevance, which is
the correct failure mode. A real fix needs a forward hook capturing the attention tensor
actually consumed inside the attention block, and we did not attempt it. We report no
faithfulness or localisation number for the method rather than a zero.

\texttt{rise} is the opposite failure and it is diagnostic. It returns a valid map on all
100 frames, yet only 4 of them carry positive energy. RISE scores random masks by the model's output and averages them. When the output is uniformly
negative on an image, no mask raises the evidence and the map has no positive part to localise. On the first-fix
bundle the detector calls 96 of those 100 images real, which is exactly the count. The 96
maps without positive energy and the 96 detector misses are the same fact seen twice, so
that edits axis is largely measuring maps of a near-constant negative output. That is a
limit of the measurement rather than a property of the methods, and it is why we
re-measured on the detector we selected. Saturation is not the whole story. On the selected detector the same method returns a valid map on all 200 frames of the unified
run. The detector fires on 75 of them and 5 maps carry positive energy, so the localisation row
of Tab.~\ref{tab:methods} rests on five frames while its AOPC advantage rests on all 200.

\section{The First-Fix Measurement}
\label{sec:supp-firstfix}

Sec.~\ref{sec:xai-results} benchmarks eighteen maps on the detector we selected. An earlier
benchmark ran the same protocol on the first-fix bundle, the champion of
Sec.~\ref{sec:shortcut}: PE-Core-336 with the forensic features on, trained on asymmetric
\texttt{webp} data. We keep that table here. The two measurements differ by detector and
not by protocol, and the difference between them is the result of Sec.~\ref{sec:xai-results}.

\begin{table*}[t]
\centering
\scriptsize
\setlength{\tabcolsep}{4pt}
\caption{The first-fix measurement: fifteen attribution methods on two axes, in the
corrected crop geometry, with 95\% cluster-bootstrap intervals. The bundle is the champion
of Sec.~\ref{sec:shortcut}, trained on asymmetric \texttt{webp} data with the forensic
features on. This is the same protocol as Tab.~\ref{tab:methods} on a different detector, and
the two tables differ by detector alone. AOPC advantage is over the \texttt{random} map,
per axis; bold marks the only intervals that exclude zero. RMA and pointing are
conditional localisation on this run's edits axis (\texttt{chefer\_relevance} produced no
valid map here, 0/100; \texttt{rise} 4/100).}
\label{tab:supp-firstfix}
\begin{tabular}{lccccc}
\toprule
Method & AOPC adv.\ edits & AOPC adv.\ gen. & RMA & Pointing & ms/map \\
\midrule
\textit{random} (control) & --- & --- & 0.179 & 0.070 & 25 \\
\textit{center} (control) & $+0.03$ {\scriptsize [$-0.50$, $+0.55$]} & $-0.11$ {\scriptsize [$-0.35$, $+0.09$]} & 0.233 & 0.680 & 23 \\
\textit{edge} (control) & $+0.22$ {\scriptsize [$-0.08$, $+0.61$]} & $-0.01$ {\scriptsize [$-0.17$, $+0.15$]} & 0.271 & 0.660 & 115 \\
\midrule
\texttt{attention\_rollout} & $+0.24$ {\scriptsize [$-0.04$, $+0.56$]} & $+0.15$ {\scriptsize [$-0.02$, $+0.31$]} & 0.205 & 0.500 & 512 \\
\texttt{raw\_attention} & $+0.07$ {\scriptsize [$-0.21$, $+0.32$]} & $+0.02$ {\scriptsize [$-0.15$, $+0.18$]} & 0.296 & 0.430 & 276 \\
\texttt{chefer\_relevance} & --- & --- & --- & --- & 104 \\
\texttt{vit\_gradcam} & $-0.17$ {\scriptsize [$-0.59$, $+0.13$]} & $-0.04$ {\scriptsize [$-0.34$, $+0.37$]} & 0.076 & 0.071 & 101 \\
\texttt{layercam} & $-0.10$ {\scriptsize [$-0.47$, $+0.21$]} & $+0.10$ {\scriptsize [$-0.07$, $+0.26$]} & 0.164 & 0.090 & 96 \\
\texttt{gradcam\_plusplus} & $-0.16$ {\scriptsize [$-0.54$, $+0.12$]} & $+0.08$ {\scriptsize [$-0.14$, $+0.40$]} & 0.185 & 0.200 & 92 \\
\texttt{integrated\_gradients} & $-0.06$ {\scriptsize [$-0.27$, $+0.13$]} & $+0.19$ {\scriptsize [$-0.02$, $+0.40$]} & 0.220 & 0.130 & 2{,}032 \\
\texttt{smoothgrad} & $-0.10$ {\scriptsize [$-0.36$, $+0.16$]} & $\mathbf{+0.14}$ {\scriptsize [$+0.00$, $+0.26$]} & 0.239 & 0.290 & 17{,}165 \\
\texttt{occlusion} & $-0.18$ {\scriptsize [$-0.57$, $+0.23$]} & $+0.25$ {\scriptsize [$-0.04$, $+0.51$]} & 0.272 & 0.530 & 35{,}080 \\
\texttt{rise} & $-0.06$ {\scriptsize [$-0.47$, $+0.32$]} & $+0.17$ {\scriptsize [$-0.06$, $+0.37$]} & --- & --- & 9{,}406 \\
\midrule
\texttt{pool\_decomposition} (ours) & $-0.27$ {\scriptsize [$-0.74$, $+0.09$]} & $\mathbf{+0.24}$ {\scriptsize [$+0.01$, $+0.51$]} & 0.228 & 0.350 & \textbf{97} \\
\texttt{pool\_occlusion} (ours) & $-0.13$ {\scriptsize [$-0.50$, $+0.16$]} & $+0.22$ {\scriptsize [$-0.03$, $+0.49$]} & 0.198 & 0.120 & \textbf{99} \\
\bottomrule
\end{tabular}
\end{table*}

\begin{figure*}[t]
  \centering
  \includegraphics[width=\textwidth]{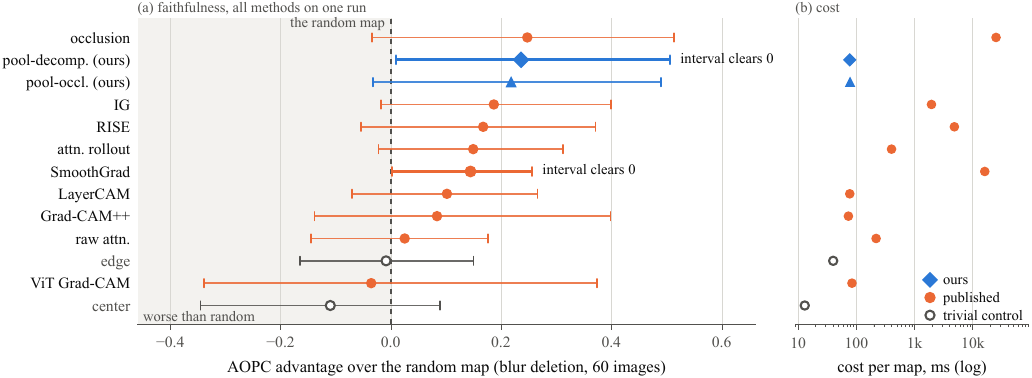}
  \caption{Faithfulness and cost on the generation axis of the first-fix bundle
  (asymmetric \texttt{webp} data, forensic features on). (a) AOPC advantage over the
  random map over 60 images, with 95\% cluster-bootstrap intervals. Rows are ordered by
  point estimate, which is not a ranking claim: only \texttt{pool\_decomposition} and
  \texttt{smoothgrad} exclude zero, and \texttt{occlusion}, which leads on the point
  estimate, does not. Two of twenty-six uncorrected tests at the 95\% level is about what
  chance produces. (b) The same rows' cost on this run, which carries no interval and so
  keeps its own panel; it does not track the advantage. Tab.~\ref{tab:methods} re-measures
  this axis on the detector we selected, where eight of seventeen maps exclude zero.}
  \label{fig:faith}
\end{figure*}

\paragraph{Why nothing cleared the control on that edits axis.}
One number bounds that run, and it is not a property of the methods.
The bundle calls 96 of the 100 edited frames real, so most of the axis measures maps of a
nearly constant negative output.
\texttt{rise} makes that visible: it returns a valid map on all 100 frames, and 96 carry
no positive energy, which are the same 96 frames the detector misses.
On the generation axis, where the detector does fire, exactly two of thirteen measurable
methods excluded zero.
The selected detector fires on 142 of the 368 edits and on 75 of the 200 frames the unified run
samples. The same protocol then separates twelve maps of seventeen from the random control on
that axis.

\paragraph{The resource limit on two gradient methods.}
The unified run shards its frames across parallel workers on one GPU.
\texttt{smoothgrad} and \texttt{integrated\_gradients} hold a batch of perturbed copies in
memory, and under that sharding they raise \texttt{CUDA out of memory} on most frames.
Their advantage was therefore first estimated on 45 of 200 edited frames and 10 of 60
generated ones.
We discarded those numbers and re-ran both methods alone, on the full samples, rather
than print a value from a truncated one.
The re-run reproduces this run's control values to three decimals, which is what lets
the two sets of rows sit in one table.
This is a limit of the machine we ran on, not a property of either method.

\begin{figure}[t]
  \centering
  \includegraphics[width=\columnwidth]{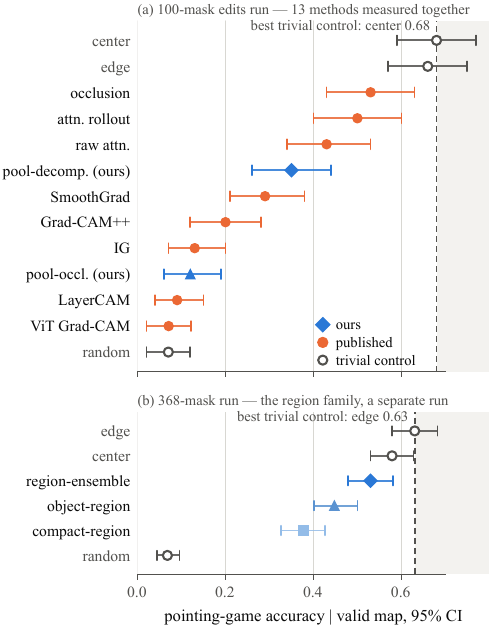}
  \caption{Pointing-game accuracy against reference intervention masks, with 95\% cluster-bootstrap intervals, on the champion bundle. Panel (b) clusters by scene, as the body does; panel (a) clusters by base image, because the scene assignment covers only the 368-mask release. (a) The 100-mask edits run, where thirteen methods were measured together. (b) The 368-mask run, which measured our region family. In both, no real method reaches the best trivial control, because edits in this set are centre-biased. The two runs disagree about which control leads, so we never merge them into one ranking and never rest a claim on one control.}
  \label{fig:loc}
\end{figure}

\begin{figure*}[t]
  \centering
  \includegraphics[width=\textwidth]{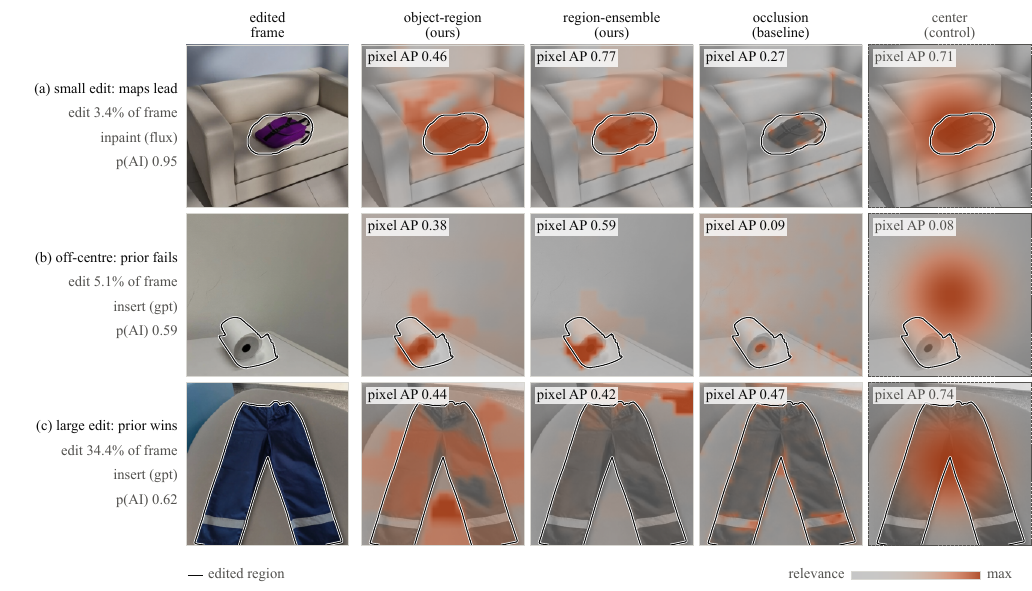}
  \caption{What the maps look like, in the corrected crop geometry. Each panel is the square crop the model sees; the outline is the reference mask and the chip is that map's pixel AP, recomputed from the array drawn. Every map is regenerated from the sealed release with the benchmark's configuration. Row (a), a 3.4\% inpaint: \texttt{region\_ensemble} (0.77) leads the \texttt{center} prior (0.71). Row (b), an off-centre insert: the prior collapses to 0.08 while our maps hold. Row (c), a 34.4\% edit: the prior wins outright, the honest failure case for a region method. Each map is scaled to its own range, so a panel shows where relevance concentrates, not how much.}
  \label{fig:qual}
\end{figure*}

\section{Region-Family Runs}
\label{sec:supp-region}

\begin{figure}[t]
  \centering
  \includegraphics[width=\columnwidth]{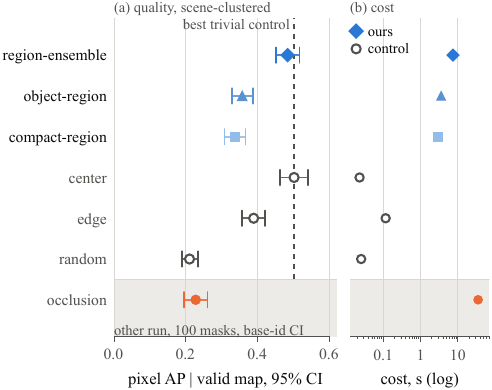}
  \caption{Region-level attribution on the 368-mask localisation run, first-fix bundle.
  (a) Quality: pixel AP with 95\% scene-clustered bootstrap intervals, every method on the
  same masks in one run. The dashed line is the best trivial control.
  \texttt{region\_ensemble} is level with it and ahead of every other real method. (b)
  Cost, which carries no interval and so keeps its own panel. We never ran \texttt{occlusion} on these masks. It sits on the shaded lane below, measured on
100 of them in the 15-method benchmark with a base-id interval, and is there for scale.}
  \label{fig:region}
\end{figure}

A separate run compares the regional method family on edit masks.
It uses the first-fix bundle of Sec.~\ref{sec:supp-firstfix}: PE-Core-336, MLP head, seed 42,
trained on asymmetric \texttt{webp} data with the forensic features on.
It covers all 368 masks, where the unified run of Tab.~\ref{tab:methods} samples 200 of them
on the detector we selected.
Fig.~\ref{fig:region} puts quality beside cost for the family. Fig.~\ref{fig:loc} puts the pointing
game of every benchmarked method beside the trivial controls, and Fig.~\ref{fig:qual} shows three
frames map by map, one per size regime.

\begin{table*}[t]
\centering
\footnotesize
\setlength{\tabcolsep}{3pt}
\caption{Region-level attribution against all 368 intervention masks in the corrected crop geometry, with 95\% cluster-bootstrap intervals over \texttt{base\_id} (2000 resamples). These are the \emph{historical} intervals; the main text prints the scene-clustered values of Tab.~\ref{tab:supp-clustering}, which change no ordering. The bundle is the champion of Sec.~\ref{sec:shortcut}, trained on asymmetric \texttt{webp} data with the forensic features on. Controls in italics. Conditional on a valid map with positive energy, as the protocol requires. $n_{\text{valid}}/N$
is 368/368 for the controls and \texttt{region\_ensemble}, 362/368 for \texttt{object\_region}
and 350/368 for \texttt{compact\_region}.}
\label{tab:supp-region}
\begin{tabular}{lccc}
\toprule
Method & pixel AP & RMA & pointing \\
\midrule
\textit{center} & \textit{0.491} {\scriptsize [0.461, 0.520]} & \textit{0.222} {\scriptsize [0.203, 0.241]} & \textit{0.579} {\scriptsize [0.530, 0.628]} \\
\textit{edge} & \textit{0.389} {\scriptsize [0.361, 0.416]} & \textit{0.261} {\scriptsize [0.241, 0.283]} & \textit{0.630} {\scriptsize [0.579, 0.682]} \\
\texttt{region\_ensemble} & \textbf{0.489} {\scriptsize [0.458, 0.519]} & 0.234 {\scriptsize [0.217, 0.253]} & 0.530 {\scriptsize [0.478, 0.582]} \\
\texttt{object\_region} & 0.358 {\scriptsize [0.330, 0.386]} & 0.306 {\scriptsize [0.280, 0.335]} & 0.448 {\scriptsize [0.401, 0.500]} \\
\texttt{compact\_region} & 0.322 {\scriptsize [0.296, 0.347]} & \textbf{0.402} {\scriptsize [0.366, 0.438]} & 0.377 {\scriptsize [0.326, 0.426]} \\
\textit{random} & \textit{0.175} {\scriptsize [0.160, 0.190]} & \textit{0.172} {\scriptsize [0.156, 0.188]} & \textit{0.068} {\scriptsize [0.043, 0.095]} \\
\bottomrule
\end{tabular}
\end{table*}

\begin{table*}[t]
\centering
\footnotesize
\setlength{\tabcolsep}{3pt}
\caption{What the geometry fix changed. Left: the map stretched over the whole frame, as in the run first reported. Right: the map placed inside the crop the model receives. Same bundle, same masks, same protocol; only the placement differs. The correction lifts the random control, and lifts the real methods more.}
\label{tab:supp-geomdelta}
\begin{tabular}{lcccccc}
\toprule
 & \multicolumn{2}{c}{pixel AP} & \multicolumn{2}{c}{RMA} & \multicolumn{2}{c}{pointing} \\
\cmidrule(lr){2-3}\cmidrule(lr){4-5}\cmidrule(lr){6-7}
Method & full frame & in crop & full frame & in crop & full frame & in crop \\
\midrule
\textit{center} & 0.497 & 0.491 & 0.171 & 0.222 & 0.576 & 0.579 \\
\textit{edge} & 0.417 & 0.389 & 0.221 & 0.261 & 0.633 & 0.630 \\
\texttt{region\_ensemble} & 0.398 & \textbf{0.489} & 0.171 & 0.234 & 0.364 & \textbf{0.530} \\
\texttt{object\_region} & 0.309 & 0.358 & 0.227 & 0.306 & 0.356 & 0.448 \\
\texttt{compact\_region} & 0.264 & 0.322 & 0.289 & \textbf{0.402} & 0.234 & 0.377 \\
\textit{random} & 0.131 & 0.175 & 0.124 & 0.172 & 0.035 & 0.068 \\
\bottomrule
\end{tabular}
\end{table*}

\begin{table*}[t]
\centering
\footnotesize
\caption{The same corrected run, end-to-end over all 368 rows: an invalid map, a map without positive energy, \emph{and a detector miss} each receive the pre-registered zero. The detector is the champion of Sec.~\ref{sec:shortcut}, trained on asymmetric \texttt{webp} data
with the forensic features on. It recognises 20 of 368 edits, so every method collapses and
none is meaningfully above the random map. The detector we selected recognises 142 of the same 368 (Tab.~\ref{tab:supp-served}), so this floor is largely a property of the bundle.}
\label{tab:supp-e2e}
\begin{tabular}{lcc}
\toprule
Method & pixel AP & pointing \\
\midrule
\textit{center} & 0.031 {\scriptsize [0.017, 0.047]} & 0.038 {\scriptsize [0.019, 0.057]} \\
\textit{edge} & 0.030 {\scriptsize [0.017, 0.044]} & 0.041 {\scriptsize [0.022, 0.062]} \\
\texttt{region\_ensemble} & 0.032 {\scriptsize [0.019, 0.047]} & 0.035 {\scriptsize [0.019, 0.054]} \\
\texttt{object\_region} & 0.029 {\scriptsize [0.017, 0.042]} & 0.035 {\scriptsize [0.019, 0.054]} \\
\texttt{compact\_region} & 0.019 {\scriptsize [0.011, 0.028]} & 0.024 {\scriptsize [0.008, 0.041]} \\
\textit{random} & 0.011 {\scriptsize [0.006, 0.018]} & 0.003 {\scriptsize [0.000, 0.008]} \\
\bottomrule
\end{tabular}
\end{table*}

Three things follow from the intervals in the corrected geometry.

First, the pixel-AP ordering among real methods is resolved, in both clusterings
(Tab.~\ref{tab:supp-clustering}).
By scene, \texttt{region\_ensemble} (0.483 [0.450, 0.516]) is clear of
\texttt{object\_region} (0.357 [0.329, 0.387]) and of \texttt{compact\_region} (0.337
[0.308, 0.366]); ensembling helps on that metric.
The ensemble is level with the \texttt{center} control (0.501 [0.461, 0.540]), so the
prior is no longer ahead on pixel AP, but neither is it beaten overall.
On relevance mass the ordering inverts, with \texttt{compact\_region} at 0.443 [0.396,
0.488] ahead of every control, and on pointing the controls still lead.
The split by edit size in Sec.~\ref{sec:supp-audit} is where the ensemble and the prior
separate.
The \texttt{base\_id} values printed in Tab.~\ref{tab:supp-region} are the historical
comparison and change none of these orderings.

Second, on this bundle \textbf{we measure no faithfulness advantage over the random map for any
method, on this axis at this protocol and sample size}.
The AOPC advantage intervals all straddle zero. \texttt{object\_region} reads $-0.755$
[$-2.590$, $0.423$] and \texttt{region\_ensemble} $-0.835$ [$-2.932$, $0.566$].
\texttt{compact\_region} reads $-1.055$ [$-3.733$, $0.674$], and even the \texttt{edge} control
$+0.204$ [$-0.103$, $0.592$].
The point estimates are negative for the region family.
The honest statement is that the edits axis has too little signal here to rank methods by
faithfulness, not that the methods are shown to be empty.
This bundle calls 96 of the 100 edited frames on the smaller axis real, so most of what
is being measured is a map of a nearly constant negative output.
On the detector we selected, on 200 of these same triples, twelve maps of seventeen clear the
random control (Tab.~\ref{tab:methods}).

Third, we read the end-to-end table (Tab.~\ref{tab:supp-e2e}) for what the conditional one hides.
With pre-registered zeros for detector misses everything drops to 0.02--0.03, because
this detector recognises only 20 of 368 local edits (5.4\%).
Reporting conditional numbers alone would misstate system quality by an order of
magnitude.
We read the collapse as a property of this bundle more than of the task. On the same masks the
detector we selected fires on 142 of 368 (Tab.~\ref{tab:supp-served}), and we did not repeat the
full method sweep on it.

\paragraph{Against the perturbation baseline with the best AOPC point estimate.}
The practical comparison is against \texttt{occlusion}.
Twenty-five generations carry both maps, though not from one run
(Tab.~\ref{tab:headtohead}).
On them \texttt{object\_region} scores higher than \texttt{occlusion} on pixel AP,
pointing and RMA, and it wins per-image on 19 of 25 frames.
The two rows sit in different crop geometries, so the size of that gap is not measured.
Cost separates the two methods further than quality does.
\texttt{object\_region} averages 3.6\,s per frame on the corrected 368-mask run, against
108\,s per \texttt{occlusion} map on the 25-frame run.
Those two means come from two runs, so we quote no within-run ratio.
We read this as a cost argument with a favourable quality sign, not as a quality claim
with an interval.
One more cross-run number belongs here.
An earlier run on the visual-only model scored \texttt{chefer\_relevance} at 0.400 on
the pointing game. On the sealed bundle of this paper it produces no valid map at all.

\paragraph{The serving code path, measured.}
The serving code path runs its own implementation of \texttt{region\_ensemble}, batched and
in processed-tensor space with 48 SLIC segments. A quantile-0.80 background floor follows,
before display.
We put that exact code path through the benchmark's localisation metrics on all 368
reference masks (Tab.~\ref{tab:supp-served}).
Two comparisons are honest, and each holds one thing fixed.
Holding the detector at the sealed bundle, the serving code path gives 0.475 pixel AP [0.445,
0.503] by \texttt{base\_id} and 0.471 [0.437, 0.505] by scene. The benchmarked implementation
on the same bundle gives 0.489 [0.458, 0.519] and 0.483 [0.450, 0.516]. The two are not
distinguishable at this sample size, which retires the caveat that we measure one artefact and
would ship another.
Holding the detector at the selected one, the raw grid gives 0.486 [0.449, 0.524] by scene. The
display variant a viewer sees gives 0.434 [0.401, 0.467], at a relevance mass of 0.476
against 0.278.
The display floor therefore trades pixel AP for concentration. Our earlier comparison of the
selected served map against the sealed benchmarked map crossed both a detector and a map
variant at once.
The two detectors differ where the maps do not.
On these 368 edits the sealed bundle fires 19 times, 5.2\% [3.0, 7.3], and the selected
detector 142 times, 38.6\% [33.7, 43.8].
The benchmark run behind Tab.~\ref{tab:supp-region} puts the sealed bundle at 20 of 368, one
frame apart from this run.

\begin{table}[t]
\centering
\footnotesize
\setlength{\tabcolsep}{4pt}
\caption{The serving implementation of \texttt{region\_ensemble} scored by the benchmark's
own localisation code against all 368 reference intervention masks, in the corrected crop
geometry. \emph{Served raw grid} is the map before the display sparsifier;
\emph{display variant} is what a viewer would see, after the quantile-0.80
background floor. Conditional on a valid map with positive energy, 367 of 368 for the raw
grid and 366 of 368 for the display variant. The two honest comparisons are within a
detector: served against benchmarked on the sealed bundle, and raw grid against display
variant on the selected one. Intervals are 95\% cluster bootstraps
over \texttt{base\_id}, except the rows marked \emph{scene}, which cluster over the set's
217 scenes (216 of which carry a valid map). \emph{Selected} is the symmetric factorial cell we chose;
\emph{sealed} is the champion of Sec.~\ref{sec:shortcut}. The last row repeats the
benchmarked implementation on the sealed bundle from Tab.~\ref{tab:supp-region}, and its
interval overlaps the served ones almost entirely.}
\label{tab:supp-served}
\begin{tabular}{llc}
\toprule
Map / detector & Metric & Value \\
\midrule
served raw grid, selected & pixel AP & 0.478 {\scriptsize [0.448, 0.509]} \\
served raw grid, selected & pixel AP (scene) & 0.486 {\scriptsize [0.449, 0.524]} \\
served raw grid, selected & patch AP & 0.476 \\
served raw grid, selected & RMA (scene) & 0.278 \\
served raw grid, selected & pointing & 0.493 \\
served raw grid, selected & IoU at top-$q$ 0.2 & 0.259 \\
served raw grid, selected & seconds / map & 5.4 \\
\midrule
display variant, selected & pixel AP & 0.435 {\scriptsize [0.404, 0.463]} \\
display variant, selected & pixel AP (scene) & 0.434 {\scriptsize [0.401, 0.467]} \\
display variant, selected & RMA (scene) & \textbf{0.476} \\
display variant, selected & pointing & 0.495 \\
\midrule
served raw grid, sealed & pixel AP & 0.475 {\scriptsize [0.445, 0.503]} \\
served raw grid, sealed & pixel AP (scene) & 0.471 {\scriptsize [0.437, 0.505]} \\
benchmarked, sealed & pixel AP & 0.489 {\scriptsize [0.458, 0.519]} \\
benchmarked, sealed & pixel AP (scene) & 0.483 {\scriptsize [0.450, 0.516]} \\
\bottomrule
\end{tabular}
\end{table}

\begin{table}[t]
  \centering\small
  \caption{\texttt{object\_region} against \texttt{occlusion}, on 25 shared generations.
\texttt{object\_region} is the component of our ensemble with the highest relevance
mass, \texttt{occlusion} the perturbation baseline with the highest AOPC point estimate. Bundle as in Tab.~\ref{tab:supp-region}. The two rows come from two runs and two geometries. We measured \texttt{occlusion} on a
25-frame run before the geometry fix and never re-measured it on these ids;
\texttt{object\_region} is the same 25 ids in the corrected geometry. The fix lifts every method
  (Tab.~\ref{tab:supp-geomdelta}), so the printed gap bounds the advantage from above rather
  than measuring it. Higher is better.}
  \label{tab:headtohead}
  \begin{tabular}{lccc}
    \toprule
    Method & pixAP & point & RMA \\
    \midrule
    \texttt{occlusion} {\scriptsize (pre-fix geometry)} & 0.204 & 0.360 & 0.197 \\
    \texttt{object\_region} {\scriptsize (corrected)} & 0.439 & 0.520 & 0.313 \\
    \bottomrule
  \end{tabular}
\end{table}

\section{The Mean-Replacement Control}
\label{sec:supp-replacement}

Sec.~\ref{sec:xai-results} reports that switching the evaluator's replacement from blur to
the channel mean reorders the table.
Tab.~\ref{tab:replacement} is that comparison.
The 120 images of the mean run are a subset of the 368. On that matched subset the blur
advantages stay negative, so the reordering is the perturbation and not the sample.

No mean-replacement interval excludes zero either.
\texttt{object\_region} gives $+2.359$ [$-0.154$, $6.561$] and \texttt{region\_ensemble}
$+1.798$ [$-0.464$, $5.421$]. \texttt{compact\_region} gives $+0.781$ [$-0.30$, $2.27$],
\texttt{occlusion} $+0.999$ [$-0.394$, $3.229$], and the \texttt{center} control $+1.412$
[$-0.250$, $3.849$].
That distribution is heavy-tailed and dominated by a handful of frames.
\texttt{object\_region}'s median advantage over \texttt{random} is $-0.077$, and it wins
on 53 of 120 images.
One paired comparison edges past zero: \texttt{object\_region} over \texttt{occlusion} at
$+1.360$ [$+0.123$, $+3.395$]. It rests on a single frame that contributes two thirds of the
paired sum. The median difference is $+0.03$, and re-clustering by scene drags the lower bound
to $+0.01$.
On this task neither replacement family separates the methods from a random map with the
evidence we have.

\begin{table}[t]
\centering
\small
\caption{AOPC advantage over the random map on the edits axis: same bundle, same maps,
evaluator's replacement switched. The bundle is the champion of Sec.~\ref{sec:shortcut},
trained on asymmetric data with the forensic features on. Blur: the 368-mask region run, in which we measured the superpixel family only, hence the dash
for \texttt{occlusion}. The 15-method run of Tab.~\ref{tab:supp-firstfix} puts it at $-0.18$
[$-0.57$, $+0.23$] under blur. Mean: 120 of
the same masks. The point
estimates reorder the table. No mean-replacement interval excludes zero, so this is a
warning about the protocol, not a new ranking.}
\label{tab:replacement}
\begin{tabular}{lcc}
\toprule
Method & blur & channel mean \\
\midrule
\texttt{object\_region}   & $-0.755$ & $+2.359$ {\scriptsize [$-0.15$, $6.56$]} \\
\texttt{region\_ensemble} & $-0.835$ & $+1.798$ {\scriptsize [$-0.46$, $5.42$]} \\
\texttt{compact\_region}  & $-1.055$ & $+0.781$ {\scriptsize [$-0.30$, $2.27$]} \\
\texttt{occlusion}        & ---      & $+0.999$ {\scriptsize [$-0.39$, $3.23$]} \\
\textit{center} (control) & $-0.761$ & $+1.412$ {\scriptsize [$-0.25$, $3.85$]} \\
\textit{edge} (control)   & $+0.204$ & $-0.659$ {\scriptsize [$-2.89$, $0.74$]} \\
\bottomrule
\end{tabular}
\end{table}

\section{Infidelity and Sensitivity in Full}
\label{sec:supp-infid}

A separate run on 184 images measures infidelity and sensitivity-max
\cite{yeh2019infidelity} for the methods that pass through the differentiable head.
In infidelity all three gradient methods are \emph{worse} than the constant controls.
\texttt{integrated\_gradients} reads 3.31, \texttt{vit\_gradcam} 2.14 and
\texttt{gradcam\_plusplus} 1.22, against ${\approx}1.1$--$1.2$ for the controls.
The perturbation effect predicted by the map departs from the actual effect more than
for a map that knows nothing about the image.
Their sensitivity-max is 0.91--1.09 against 0.00 for the controls.
Input noise of $\sigma = 0.03$ reorders pixel importance almost arbitrarily, which
reproduces the known fragility of explanations
\cite{ghorbani2019fragile,kindermans2019unreliability} on an applied task.
The zero of the controls is no virtue, since a constant map is trivially stable.
The stability block of the same benchmark puts \texttt{attention\_rollout} at $\rho = 0.9956$
on the edits axis and $0.9940$ on the generation axis. That makes it the most stable method we
measured.

\section{The Localisation Geometry Defect}
\label{sec:supp-geom}

The processor resizes the shorter edge to 336 and centre-crops a square.
On a non-square frame the model therefore receives only the central band.
The repository's own helper reproduces that crop, and the map overlay uses it.
The metric path did not: it upsampled the $24\times24$ map to the full frame instead.
Recomputing the two analytic controls over all 368 masks in both geometries isolates
the effect.
Those two controls can be regenerated exactly without the model.

\begin{center}
\footnotesize
\begin{tabular}{llccc}
\toprule
Control & Metric & full frame & inside crop & $\Delta$ \\
\midrule
\textit{center} & pixel AP & 0.4971 & 0.4914 & $-0.006$ \\
\textit{center} & RMA      & 0.1711 & 0.2217 & $+0.051$ \\
\textit{center} & pointing & 0.5761 & 0.5788 & $+0.003$ \\
\textit{random} & pixel AP & 0.1311 & 0.1752 & $+0.044$ \\
\textit{random} & RMA      & 0.1245 & 0.1715 & $+0.047$ \\
\textit{random} & pointing & 0.0353 & 0.0679 & $+0.033$ \\
\bottomrule
\end{tabular}
\end{center}

The full-frame column reproduces the published run to four decimals.
That is how we identify the geometry we computed the published numbers in.

The fix threads the crop box into every localisation quantity, not only the pixel map.
Relevance outside the visible band now scores zero.
The patch-level metric compares the map's grid against the mask \emph{restricted to the
crop} rather than against the whole frame.
Previously a target lying entirely in the cropped-away band scored a chance-level 0.5
instead of being undefined.
We evaluate the sub-patch-grid stress flag on the grid the model actually had.
We also report and aggregate the share of the target falling outside the band: 98 of
the 368 masks are partly outside it, 5.1\% of mask area on average.
Nine tests cover the three failure modes and the runner-level path, and we pin the
pre-fix behaviour in the suite so the regression stays visible.
We measured every localisation number in this paper after the fix, with one marked exception.
The \texttt{occlusion} row of Tab.~\ref{tab:headtohead} comes from a 25-frame run that
predates the fix, and we never repeated that run on those ids.

\begin{figure*}[t]
  \centering
  \includegraphics[width=\textwidth]{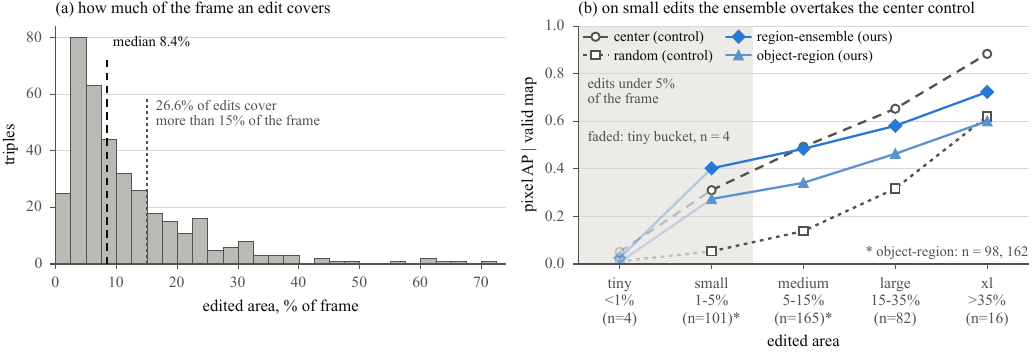}
  \caption{Audit of the mask set: all 368 triples, recomputed from the images themselves. (a) The edited area has median 8.4\% of the frame, but 26.6\% of edits cover more than 15\% and the largest covers 71\%, so ``local edit'' spans two orders of magnitude. (b) Pixel AP by the protocol's pre-registered size buckets, on the champion bundle. Every method rises with area, and the \texttt{center} control rises fastest, so its overall lead is largely an artefact of large centred edits. On the small bucket \texttt{region\_ensemble} overtakes it: $+0.098$ paired advantage over 71 scenes, scene-clustered 95\% CI $[+0.026, +0.169]$.}
  \label{fig:maskaudit}
\end{figure*}

\section{Audit of the Mask Set}
\label{sec:supp-audit}

All numbers in this section are recomputed from the images themselves plus the run's
stored per-image localisation records, and Tab.~\ref{tab:supp-audit} is the summary.
The seven scripts that produce them are released with the paper sources.

\begin{table}[t]
\centering
\footnotesize
\caption{The 368 triples, measured rather than asserted. Mean absolute difference is per
channel on a 0--255 scale, between the background frame and the edited frame.}
\label{tab:supp-audit}
\begin{tabular}{lr}
\toprule
Quantity & Value \\
\midrule
Triples / distinct scenes & 368 / 217 \\
Backgrounds near-duplicated at least once & 235 (63.9\%) \\
Largest duplicate group & 7 triples \\
\midrule
Difference inside the mask (median) & 59.0 \\
Difference outside the mask (median) & 1.58 \\
Difference outside the mask (p90 / max) & 3.62 / 31.3 \\
Triples with outside difference $>5$ & 15 (4.1\%) \\
\midrule
Edited area: median / mean & 8.4\% / 12.1\% \\
Edited area: p05 / p95 / max & 2.0\% / 32.5\% / 71.0\% \\
Edits below 1\% / above 15\% of the frame & 4 / 98 \\
\bottomrule
\end{tabular}
\end{table}

Three consequences follow, all in the corrected crop geometry.

\emph{(i)} Duplicate backgrounds mean a cluster bootstrap over \texttt{base\_id} treats 368
units where 217 exist. The scene is therefore the primary bootstrap unit of this paper, and we
keep \texttt{base\_id} only as a comparison against the historical numbers
(Tab.~\ref{tab:supp-clustering}).
Across methods and metrics, re-clustering by scene widens the conditional intervals by
\textbf{4.8--63.7\%}: least for \texttt{object\_region} pixel AP (4.8\%) and most for
\texttt{random} pointing (63.7\%).
Every ordering stays intact, and the ensemble stays level with the \texttt{center}
control under either clustering.
Exactly one significance verdict changes anywhere in the recomputation.
\texttt{compact\_region}'s paired deficit against the centre prior on the \texttt{small} bucket
reads $-0.082$ [$-0.147$, $-0.017$] by \texttt{base\_id}. By scene it reads $-0.074$ [$-0.150$,
$+0.001$], and stops excluding zero.

\emph{(ii)} Strict composition is violated in 15 triples, which are whole-frame
regenerations rather than inpaints.
Excluding them changes no conclusion.
Recomputing what each tool actually changed, we also see how far an edit escapes the requested
mask. The median triple leaks 17.5\% of its changed area outside it, and 58.4\% of triples leak
more than 5\%.
That pooled median hides a bimodal split by generator.
Median leakage is 34.5\% for the 183 \texttt{gpt} triples, 1.14\% for the 135 \texttt{flux}
triples and 0.19\% for the 50 \texttt{bria} triples. In 37 triples more of the realised change
falls outside the requested mask than inside.
A requested mask is therefore a usable target for some tools and a loose one for others. The
generator composition of a size bucket matters too: 78 of the 101 triples in the pre-registered
\texttt{small} bucket are \texttt{gpt}, against 15 \texttt{flux} and 8 \texttt{bria}.

\emph{(iii)} The edited area spans two orders of magnitude, and the centre control's
lead is a function of it (Fig.~\ref{fig:maskaudit}b).
Its pixel AP is 0.310 on the \texttt{small} bucket against 0.884 on \texttt{xl}.
Bucket occupancy is 4 / 101 / 165 / 82 / 16 for tiny / small / medium / large / xl.
Per pre-registered bucket, we compare \texttt{region\_ensemble} against \texttt{center}. The
scene-clustered paired advantage is $-0.025$ [$-0.074$, $+0.011$] on \texttt{tiny} (4 scenes)
and $+0.098$ [$+0.026$, $+0.169$] on \texttt{small} (71). The historical \texttt{base\_id}
figures for the same two buckets are $-0.025$ [$-0.074$, $+0.011$] and $+0.092$ [$+0.028$,
$+0.159$], over 4 and 101 triples.
The point estimate then turns negative. It reads $-0.031$ [$-0.087$, $+0.028$] on
\texttt{medium} (113), $-0.072$ [$-0.125$, $-0.023$] on \texttt{large} (80) and $-0.160$
[$-0.238$, $-0.093$] on \texttt{xl} (16).
The sign flips once above the \texttt{tiny} bucket, whose 4 scenes resolve nothing, at
the small-to-medium boundary.
This is why we report the small-edit regime separately.
It is both the harder and the more realistic one, and the only regime in which a real
method beats the strongest trivial control on pixel AP.

\begin{table*}[t]
\centering
\scriptsize
\setlength{\tabcolsep}{3pt}
\caption{Both clusterings of the 368-mask localisation run, conditional, in the corrected
crop geometry. \emph{base} is the historical cluster bootstrap over \texttt{base\_id} (368 units).
\emph{scene} is the primary one used throughout the main text, at 217 units --- 216 for
\texttt{object\_region} and 203 for \texttt{compact\_region}, which do not return a valid map
on every triple. 2000 resamples, 95\% intervals. No ordering changes. Across
all methods and metrics the interval widens by 4.8--63.7\%.}
\label{tab:supp-clustering}
\begin{tabular}{lcccccc}
\toprule
 & \multicolumn{2}{c}{pixel AP} & \multicolumn{2}{c}{RMA} & \multicolumn{2}{c}{pointing} \\
\cmidrule(lr){2-3}\cmidrule(lr){4-5}\cmidrule(lr){6-7}
Method & base & scene & base & scene & base & scene \\
\midrule
\textit{center} & \textit{0.491} {\scriptsize [0.461, 0.520]} & \textit{0.501} {\scriptsize [0.461, 0.540]} & \textit{0.222} {\scriptsize [0.203, 0.241]} & \textit{0.262} {\scriptsize [0.236, 0.289]} & \textit{0.579} {\scriptsize [0.530, 0.628]} & \textit{0.585} {\scriptsize [0.525, 0.646]} \\
\textit{edge} & \textit{0.389} {\scriptsize [0.361, 0.416]} & \textit{0.388} {\scriptsize [0.355, 0.420]} & \textit{0.261} {\scriptsize [0.241, 0.283]} & \textit{0.285} {\scriptsize [0.257, 0.312]} & \textit{0.630} {\scriptsize [0.579, 0.682]} & \textit{0.635} {\scriptsize [0.576, 0.692]} \\
\texttt{region\_ensemble} & 0.489 {\scriptsize [0.458, 0.519]} & 0.483 {\scriptsize [0.450, 0.516]} & 0.234 {\scriptsize [0.217, 0.253]} & 0.268 {\scriptsize [0.243, 0.294]} & 0.530 {\scriptsize [0.478, 0.582]} & 0.528 {\scriptsize [0.470, 0.588]} \\
\texttt{object\_region} & 0.358 {\scriptsize [0.330, 0.386]} & 0.357 {\scriptsize [0.329, 0.387]} & 0.306 {\scriptsize [0.280, 0.335]} & 0.321 {\scriptsize [0.288, 0.356]} & 0.448 {\scriptsize [0.401, 0.500]} & 0.446 {\scriptsize [0.387, 0.507]} \\
\texttt{compact\_region} & 0.322 {\scriptsize [0.296, 0.347]} & 0.337 {\scriptsize [0.308, 0.366]} & 0.402 {\scriptsize [0.366, 0.438]} & 0.443 {\scriptsize [0.396, 0.488]} & 0.377 {\scriptsize [0.326, 0.426]} & 0.427 {\scriptsize [0.363, 0.488]} \\
\textit{random} & \textit{0.175} {\scriptsize [0.160, 0.190]} & \textit{0.210} {\scriptsize [0.188, 0.233]} & \textit{0.172} {\scriptsize [0.156, 0.188]} & \textit{0.208} {\scriptsize [0.186, 0.232]} & \textit{0.068} {\scriptsize [0.043, 0.095]} & \textit{0.112} {\scriptsize [0.071, 0.155]} \\
\bottomrule
\end{tabular}
\end{table*}

A rebuild of the set to the original specification remains future work: held-out
provenance, three mask scales per base image, lossless storage.
It requires paid generation.
The audit above bounds how much the present set can be distorting the conclusions.

\section{Provenance of Every Figure and Table}
\label{sec:supp-provenance}

The material below is scattered through the paper by design, since every caption names
its own bundle. We collect it in one place here. For each numbered item we give the data we computed it on, the
model bundle behind it, the run that produced it, and the configuration and seed. Dataset ids are SHA-256 of content. Bundle ids are content
hashes of the frozen release. Run ids are the cluster job uuids, truncated to eight
characters, or the script for work done outside the scheduler. A dash means the quantity
does not apply, for example a chart with no model in it.

\begin{table*}[t]
\centering
\scriptsize
\setlength{\tabcolsep}{3pt}
\caption{Figure and table provenance. \emph{sealed} $=$ \texttt{1b2ed107\ldots}, the
champion: PE-Core-336, forensic features on, trained on \texttt{dataset-v2-train-webp}
(\texttt{b94c9140\ldots}) in run \texttt{990ec15b} at seed 42. \emph{selected} $=$ \texttt{2205f1e7\ldots}, the winning factorial cell: PE-Core-336, forensic
features on, trained on \texttt{dataset-v2-train-webp2} (\texttt{3153e678\ldots}) in run
\texttt{c0a9c047}, at threshold 0.6443. Its provenance is git \texttt{76fb4a22}, container
\texttt{pytorch/pytorch@sha256:c8268a92\ldots}, lock \texttt{ccce4af3\ldots}.
\emph{visual-only} $=$ \texttt{4104200b}, the visual-only model on the same
\texttt{webp2} data. \emph{xaiset} $=$ \texttt{dataset-v2-test-xai-webp}, the 368
triples.}
\label{tab:supp-provenance}
\begin{tabular}{lllll}
\toprule
Item & Data & Bundle & Run & Config / seed \\
\midrule
\multicolumn{5}{l}{\emph{Main text}} \\
Fig.~\ref{fig:teaser} & xaiset, 1 frame & selected & \texttt{eval\_served\_maps.py} & \texttt{region\_ensemble} display, 48 seg. \\
Fig.~\ref{fig:axes} & xaiset, 200 masks / 60 gen. & selected & \texttt{lane\_unified}, \texttt{lane\_gen} & same rows both panels, sample seed 42 \\
Tab.~\ref{tab:factorial} & \texttt{-webp}/\texttt{-webp2}; native test & 12 cells & \texttt{seedstudy\_factorial}, \texttt{c0a9c047} & seeds 42/43/44, 3 per cell \\
Tab.~\ref{tab:methods} & xaiset, 200 masks / 60 gen. & selected & \texttt{lane\_unified}, \texttt{lane\_gen} & blur, corrected geometry, sample seed 42 \\
\midrule
\multicolumn{5}{l}{\emph{Appendix}} \\
Fig.~\ref{fig:profile} & \texttt{dataset-v2-\{train,test\}} & \texttt{afc3f7ec}, \texttt{f15b0cba}; 5 public & \texttt{7cd86b6e} & full features, seed 42; no revision pinned \\
Tab.~\ref{tab:supp-generators} & \texttt{dataset-v2} manifest & --- & collector & --- \\
Tab.~\ref{tab:supp-splits} & \texttt{11876f70\ldots}, \texttt{8434ebf1\ldots} & --- & splitter & \texttt{StratifiedGroupKFold}(5), fold 0 \\
Tab.~\ref{tab:supp-pergen} & \texttt{dataset-v2-train} (native) & \texttt{afc3f7ec}, \texttt{f15b0cba} & 2026-07-20, 2026-07-21 & full features, seed 42 \\
Tab.~\ref{tab:supp-channels} & \texttt{3153e678\ldots}, 42{,}991 images & one PE-Core-336 pass & probes \texttt{P1}--\texttt{P3} & $d\to256\to64\to1$, seed 42 \\
Tab.~\ref{tab:supp-history} & native test, 1{,}500 $+$ 1{,}500 & sealed, selected & \texttt{eval\_codec\_invariance.py} & 8 histories, sampling seed 0 \\
Tab.~\ref{tab:supp-logo} & \texttt{-train-webp2} / \texttt{-test-webp2} & per fold & \texttt{e64a47d2}, \texttt{0a680ffc}, \texttt{ec3be298} & visual-only, one fold and one seed each \\
Tab.~\ref{tab:detector} & \texttt{-webp} vs \texttt{-webp2}; native test & sealed, visual-only & \texttt{990ec15b}, \texttt{c0a9c047} & seed 42 \\
Tab.~\ref{tab:supp-firstfix} & xaiset, 100 masks / 60 gen. & sealed & \texttt{5deec0f5}, \texttt{0440ffef} & blur, corrected geometry, $n_{\text{boot}}=2000$ \\
Fig.~\ref{fig:faith} & xaiset, 60 generations & sealed & \texttt{0440ffef} & blur, corrected geometry \\
Fig.~\ref{fig:region} & xaiset, 368 masks $+$ 25-frame pilot & sealed & \texttt{4d6a84ab}; pilot pre-fix & blur; pilot in pre-fix geometry \\
Tab.~\ref{tab:supp-region} & xaiset, 368 masks & sealed & \texttt{4d6a84ab} & scene and \texttt{base\_id} bootstraps \\
Tab.~\ref{tab:supp-clustering} & xaiset, 368 masks & sealed & \texttt{4d6a84ab} $+$ \texttt{recompute.py} & 2000 resamples, bootstrap seed 0 \\
Tab.~\ref{tab:supp-geomdelta} & xaiset, 368 masks & sealed & \texttt{7ecea406} vs \texttt{4d6a84ab} & full-frame vs in-crop map placement \\
Tab.~\ref{tab:supp-e2e} & xaiset, 368 masks & sealed & \texttt{4d6a84ab} & pre-registered zeros for misses \\
Tab.~\ref{tab:supp-served} & xaiset, 368 masks & selected, sealed & \texttt{eval\_served\_maps.py} & 48 seg., $q_{0.80}$ floor, bootstrap seed 0 \\
Tab.~\ref{tab:headtohead} & xaiset, 25 shared generations & sealed & \texttt{4d6a84ab}; pilot pre-fix & two runs, two geometries \\
Tab.~\ref{tab:replacement} & xaiset, 368 and 120 masks & sealed & \texttt{4d6a84ab}, \texttt{483609f4} & blur vs channel-mean replacement \\
Tab.~\ref{tab:supp-protocol} & --- & --- & benchmark config & as released with the sources \\
Tab.~\ref{tab:supp-methods} & --- & --- & registry & 28 implementations \\
Tab.~\ref{tab:supp-audit} & xaiset, 368 triples & --- & \texttt{audit\_masks.py}, \texttt{realized\_mask\_audit.py} & $\tau = \max(3, 1.5\,q_{0.999})$ \\
Fig.~\ref{fig:tradeoff} & \texttt{-webp} vs \texttt{-webp2}; native test & sealed, visual-only & \texttt{990ec15b}, \texttt{c0a9c047} & seed 42 \\
Fig.~\ref{fig:loc} & xaiset, 100 masks $+$ 368 masks & sealed & \texttt{5deec0f5}, \texttt{4d6a84ab} & two runs, tagged in the chart \\
Fig.~\ref{fig:qual} & xaiset, 3 frames & sealed & regenerated from the sealed release & benchmark configuration \\
Fig.~\ref{fig:maskaudit} & xaiset, 368 triples & sealed (panel b) & \texttt{fig\_data.py}, \texttt{4d6a84ab} & pre-registered size buckets \\
\bottomrule
\end{tabular}
\end{table*}

Three caveats travel with this table.
We fetched the five public detectors of Fig.~\ref{fig:profile} from their Hugging Face default
branches during run \texttt{7cd86b6e}, without a pinned revision. That row is therefore
reproducible in procedure and not in bytes.
We trained the final model outside the scheduled loop, under a lock file that differs from the
one its bundle carries (\texttt{transformers} 4.48.3 against 4.57.6). The bundle verifier
therefore checks internal consistency rather than execution.
And raw per-run artefacts live in the cluster's artifact store rather than in this repository.
A reader can therefore verify the procedure from source, but cannot re-derive a specific number
without those artefacts.

\end{document}